\documentclass[lettersize,journal]{IEEEtran}

\usepackage{amsmath,amsfonts}
\usepackage{amssymb}
\usepackage{algorithm}
\usepackage[noend]{algpseudocode}
\usepackage{array}
\usepackage[caption=false,font=normalsize,labelfont=sf,textfont=sf]{subfig}
\usepackage{textcomp}
\usepackage{stfloats}
\usepackage{url}
\usepackage{verbatim}
\usepackage{graphicx}
\usepackage{cite}

\usepackage{cmap}
\usepackage[T1]{fontenc}
\usepackage{color, colortbl}
\usepackage{multirow}
\usepackage{makecell}
\usepackage{booktabs}
\usepackage{tikz}
\usetikzlibrary{arrows.meta}
\usepackage{pgfplots}
\pgfplotsset{compat=1.18}
\usepackage{float}
\usepackage{comment}
\usepackage[hidelinks]{hyperref}

\definecolor{LightCyan}{rgb}{0.8,0.9,1}

\let\origsection\section
\let\origsubsection\subsection
\let\origsubsubsection\subsubsection
\let\origparagraph\paragraph

\let\chapter\origsection
\let\section\origsubsection
\let\subsection\origsubsubsection
\let\subsubsection\origparagraph

\let\origtable\table
\let\endorigtable\endtable
\let\origfigure\figure
\let\endorigfigure\endfigure

\renewenvironment{figure}[1][!t]
  {\begin{figure*}[#1]}
  {\end{figure*}}

\renewenvironment{table}[1][!t]
  {\begin{table*}[#1]}
  {\end{table*}}

\newenvironment{narrowtable}[1][!t]{\origtable[#1]}{\endorigtable}
\newenvironment{narrowfigure}[1][!t]{\origfigure[#1]}{\endorigfigure}

\providecommand{\afterpage}[1]{}

\begin{document}

\title{Efficient PEFT Methods with Adaptive Checkpointing for Vision Models and VLMs on Resource Constrained Consumer-GPUs}
%

\author{Altay~Toktassyn~and~Jurn-Gyu~Park%
\thanks{A. Toktassyn and J. G. Park are with the Department of Computer Science,
Nazarbayev University, Astana 010000, Kazakhstan
(e-mail: altay.toktassyn@nu.edu.kz; jurn.park@nu.edu.kz).}%
\thanks{Source code and reproducibility scripts are available at
\url{https://github.com/JokerOneK/cifar100_vits16_repro}.}}

\markboth{Preprint,~2026}%
{Toktassyn \MakeLowercase{\textit{and}} Park: Efficient PEFT Methods with Adaptive Checkpointing for Vision Models}

\maketitle

\IEEEpeerreviewmaketitle


{
\begin{abstract}
Modern pretrained vision models achieve strong accuracy but demand substantial GPU memory for fine-tuning, making edge deployment impractical. This paper compares five parameter-efficient fine-tuning (PEFT) methods (Full FT, LoRA, AdaLoRA, QLoRA, BitFit) on Transformers- (ViT-Small, TinyViT) and Mamba-based vision backbones (Vim-Small, MambaVision-T) under an on-device VRAM budget (e.g., 2\,GB), together with three gradient-checkpointing strategies (none, static, and a proposed memory-budget-aware adaptive algorithm); and we evaluate three families of foundation-model baselines: zero-shot contrastive vision language models (OpenCLIP, SigLIP), self-supervised vision backbones with lightweight evaluation protocols (DINOv2), and autoregressive VLMs for prompt-based classification (PaliGemma, MobileVLM, SmolVLM). 
%
Experiments on CIFAR-100 and DTD report accuracy, training time, energy, and the NetScore family of multi-objective metrics, which we extend with two deployment-aware variants. QLoRA and BitFit cut energy 20--30\,\% at a 1--2\,\% accuracy cost; the adaptive algorithm reduces peak memory 43--79\,\% with 9--30\,\% energy overhead. DINOv2 surpasses fine-tuned models on CIFAR-100 (0.917 vs.\ 0.897) at a fraction of the energy, while small autoregressive VLMs remain uncompetitive.
\end{abstract}

}



{%
\chapter{Introduction}
\label{ch:introduction}


{

Deep-learning(DL) models for computer vision (CV) have improved significantly in recent years both in parameter size and performance. Initially, it started from convolutional neural networks (CNNs) with a small number of parameters. They quickly evolved into Vision Transformers (ViTs)~\cite{dosovitskiy2020image}, hybrid CNN+ViT architectures such as TinyViT~\cite{wu2022tinyvit}, Mamba-based vision backbones~\cite{gu2023mamba} such as Vim-Small~\cite{zhu2024vision}, and Mamba+ViT hybrids like MambaVision-T~\cite{hatamizadeh2024mambavision}.
%
However, their main disadvantage is their need for large resources. Even the smaller variants in a model family (e.g., ViT-Small with 22M parameters, as opposed to ViT-Large with 307M) consume a substantial amount of GPU memory (VRAM) and system RAM just to store the pretrained weights---and fine-tuning further inflates this budget via optimizer states, gradients, and activations. 
This creates a big problem for users with limited budgets who do not have access to powerful cloud GPUs.
This issue is critical today: many users who practice deep learning for personal purposes can only use the limited capabilities of their hardware. For example, edge-AI devices such as Nvidia Jetson Nano provide only 2--8\,GB of system memory, and consumer-grade GPUs (e.g., GTX~1650 with 4\,GB VRAM) offer similarly tight budgets. 
Under such constraints, it is very difficult to fine-tune modern vision and vision-language models; and we cannot simply store all the activation tensors in memory during the forward/backward pass. 

Although modern vision models achieve strong accuracy, fine-tuning them on resource-constrained devices faces two fundamental challenges.
First, \emph{memory}: fine-tuning requires storing model weights, optimizer states, gradients, and intermediate activations simultaneously; on devices with 2--8\,GB VRAM, even models with 20--30M parameters frequently trigger out-of-memory (OOM) failures.
Second, \emph{the activation bottleneck}: reducing the number of trainable parameters does not proportionally reduce memory usage, because activation tensors from the forward pass---not the weights themselves---dominate GPU memory during training. This means that parameter-efficient approaches alone cannot guarantee that fine-tuning fits within a tight memory budget without additional memory-management techniques. 

Moreover, fine-tuning DL models on edge devices is constrained by strict power budgets. For example: edge platforms such as Nvidia Jetson Nano operate at 5--10\,W~\cite{nvidia_jetson_nano}, and even consumer GPUs like the GTX 1650 goes up to 75\,W~\cite{nvidia_gtx1650}. These numbers are far below the 300--700\,W envelopes of data-center accelerators~\cite{patterson2021carbon,nvidia_h100}. 
Under such limits, on-device fine-tuning can quickly exhaust battery reserves or trigger thermal throttling, making energy a first-order design constraint alongside accuracy. Schwartz et al.~\cite{schwartz2020green} formalized this concern through the \emph{Green AI} initiative, advocating that computational efficiency should be a primary evaluation metric. 
On on-device fine-tuning platforms, this principle becomes a hard constraint rather than a guideline; and the choice of model architecture, PEFT method, and checkpointing strategy directly determines whether fine-tuning can complete within the energy and thermal budget of the target hardware.




To resolve these challenges, parameter-efficient fine-tuning (PEFT) methods---such as LoRA~\cite{hu2021lora}, AdaLoRA~\cite{zhang2023adalora}, QLoRA~\cite{dettmers2023qlora}, and BitFit~\cite{benzaken2021bitfit}---tackle the memory problem from one side, freezing most of the pretrained weights and train only a small subset of parameters; and static gradient checkpointing~\cite{chen2016training} attacks it from another, discarding activations during the forward pass and recomputed during backpropagation, and trading extra compute for lower memory use.
%
While PEFT and static checkpointing alleviate on-device constraints, they are applied uniformly without regard to runtime memory, latency, and energy availability. 
Therefore, we compare Transformer and Mamba architectures under the same PEFT configuration, since model architecture fundamentally affects memory behavior: Mamba models maintain fixed memory per token (i.e., linear scale), unlike Transformers whose attention mechanism scales quadratically.
We further propose adaptive gradient checkpointing that monitors GPU memory at runtime and selectively recomputes only the layers exceeding a threshold, which helps to avoid unneeded overhead when memory is available. 
In addition to PEFT-based fine-tuning, we evaluate three families of foundation-model baselines via contrastive VLMs (OpenCLIP, SigLIP), a knowledge-distillation self-supervised model (DINOv2), and autoregressive VLMs (PaliGemma, MobileVLM, SmolVLM), establishing whether task-specific fine-tuning is justified over training-free alternatives under the same energy constraints. 
}

{
Our contributions of the paper are as follows:

\begin{itemize}
\item
We evaluate five PEFT methods across Transformer (ViT-S, TinyViT) and Mamba (Vim-S, MambaVision-T) backbones, ranking every (architecture, PEFT) pair, using quantitative metrics. 

\item We propose two quantitative multi-objective NetScore variants: \textit{NS$_{M}$} for accuracy-memory trade-offs and \textit{NS$^{\#}$} for accuracy-memory-energy trade-offs, in addition to adopting \textit{NS} and \textit{NS$_{E}$}. 


\item
We propose a memory-budget-aware adaptive checkpointing method, monitoring GPU memory at runtime and recomputing only the layers exceeding a budget-derived threshold. 

\item
We compare vision models against three families of foundation-model baselines (contrastive, self-supervised vision backbone, and autoregressive VLMs) in both accuracy and energy. 
%

\end{itemize}



The remainder of this paper is organized as follows: 
Section~\ref{ch:background} introduces the problem statement, formulates the questions, and surveys related work. Section~\ref{ch:methodology} describes the methodology---PEFT methods, adaptive checkpointing, evaluation metrics, experimental setup including datasets, hardware, and training settings. Section~\ref{ch:results} presents our results and analysis. Section~\ref{sec:discussion} presents the discussion. Section~\ref{ch:conclusion} is the conclusion of the paper.
}



{
\chapter{Motivation and Related Work}
\label{ch:background}


\section{Motivation}
\label{sec:research-gap}

\textbf{Problem Statement}: Three open problems motivate this paper.
\emph{First}, most PEFT benchmarks rank methods on accuracy and trainable-parameter count alone, ignoring the memory, energy, and latency that determine whether a (PEFT, architecture) pair is actually deployable on consumer-grade hardware; and they cover Transformer backbones only, leaving the Transformer-vs.-Mamba trade-off under PEFT unexplored (Q1).
\emph{Second}, gradient checkpointing is applied uniformly --- either off or per-layer --- with no mechanism that decides per layer whether recomputation is actually necessary at the current runtime VRAM, leaving avoidable energy and time overhead on the table (Q2).
\emph{Third}, the rapid progress of training-free foundation-model baselines --- contrastive VLMs (OpenCLIP, SigLIP), knowledge-distillation self-supervised (KD-SSL) backbones (DINOv2), and autoregressive VLMs (PaliGemma, MobileVLM, SmolVLM) --- has reopened the question of whether task-specific fine-tuning is still justified under matched accuracy--energy budgets, especially on fine-grained domains (Q3).
}



{
Therefore, this paper addresses three questions to resolve the problems: 

\textit{Q1: Which combination of PEFT method and architecture family (Transformer vs.\ Mamba) achieves the best efficiency, and  what kinds of quantitative metrics can be utilized? 
}
%

\textit{Q2: How does adaptive gradient checkpointing compare to the default checkpointing in terms of memory- and energy-efficiency?}

\textit{Q3: Does PEFT fine-tuning remain competitive against zero-shot VLM baselines in terms of accuracy-energy tradeoffs?}
}




{
\section{Related Work}
\label{sec:related-summary}


\textbf{Vision Transformers and Mamba}: Dosovitskiy et al.~\cite{dosovitskiy2020image} introduced ViT and used the Transformer architecture~\cite{vaswani2017attention} directly on image patches for classification, affecting additional ViT variants of DeiT~\cite{touvron2021training} and Swin Transformer~\cite{liu2021swin}.
%
%
Wu et al.~\cite{wu2022tinyvit} developed TinyViT, a compact hybrid that combines MBConv and attention blocks using fast pretraining distillation. 
Gu et al.~\cite{gu2022efficiently} proposed S4, a structured state space model that is effective at modeling long-range sequences. This model was later refined with selective state spaces and linear-time complexity and named Mamba~\cite{gu2023mamba}.
Zhu et al.~\cite{zhu2024vision} adapted the Mamba model to the vision of the task and named it Vim, which was made possible by bidirectional scanning of image patch sequences.
Hatamizadeh and Kautz~\cite{hatamizadeh2024mambavision} presented MambaVision, a hybrid architecture that combined Mamba blocks and Transformer attention blocks.

\textbf{Parameter-Efficient Fine-Tuning}: LoRA~\cite{hu2021lora} injects trainable low-rank matrices into frozen weight layers, reducing trainable parameters to under 1\% while preserving most of the full fine-tuning accuracy.
AdaLoRA~\cite{zhang2023adalora} extends LoRA with adaptive rank allocation that distributes a global rank budget across layers based on learned importance scores.
QLoRA~\cite{dettmers2023qlora} combines 4-bit NormalFloat quantization of frozen weights with LoRA adapters, enabling fine-tuning of large models on consumer GPUs.
BitFit~\cite{benzaken2021bitfit} trains only bias terms ($<$0.1\% of parameters) and establishes a lower bound on what minimal adaptation can achieve.
Broader surveys of PEFT~\cite{houlsby2019adapters, pfeiffer2020adapterhub, han2024peft} catalog the full landscape of adapters, prefix-tuning, and prompt-tuning variants.
All the methods were originally proposed and evaluated in NLP settings; none report energy consumption or operate under strict VRAM constraints for vision tasks.

\textbf{Activation Checkpointing}: Chen et al.~\cite{chen2016training} proposed gradient checkpointing, trading recomputation for memory by discarding intermediate activations during the forward pass and recomputing them during backpropagation.
Kirisame et al.~\cite{kirisame2020dynamic} introduced dynamic tensor rematerialization (DTR), a runtime heuristic that evicts and recomputes tensors based on cost--memory trade-offs.
Jain et al.~\cite{jain2020checkmate} formulated optimal checkpointing as an integer linear program (ILP), finding the minimum-cost recomputation schedule for a given memory budget.
Both DTR and Checkmate target full model training rather than PEFT, where the memory dynamics differ substantially due to frozen weights and adapter-only gradients.

\begin{table}[htbp]
\centering
\caption{Comparison of this study with related work across key dimensions.}
\label{tab:related-work}
\small
\begin{tabular}{lcccccc}
\hline
\textbf{Study} & \textbf{\shortstack{Multiple\\PEFT}} & \textbf{\shortstack{Vision\\Mamba}} & \textbf{\shortstack{Energy\\Measured}} & \textbf{\shortstack{Adaptive\\Ckpt.}} & \textbf{\shortstack{Efficiency\\Metrics}} & \textbf{\shortstack{Memory\\Constraint}} \\
\hline
LoRA~\cite{hu2021lora}             & No      & No  & No  & No  & No      & No       \\ 
AdaLoRA~\cite{zhang2023adalora}     & Partial & No  & No  & No  & No      & No       \\
QLoRA~\cite{dettmers2023qlora} & Partial & No  & No  & No  & No      & Yes      \\ 
Vim~\cite{zhu2024vision}          & No      & Yes & No  & No  & No      & No       \\
MambaVision~\cite{hatamizadeh2024mambavision} & No & Yes & No & No & No & No \\ 
DTR~\cite{kirisame2020dynamic} & No    & No  & No  & Yes & No      & No       \\ 
Checkmate~\cite{jain2020checkmate}     & No      & No  & No  & Yes & No      & Yes      \\ 
Green AI~\cite{schwartz2020green} & No      & No  & Yes & No  & Partial & No       \\ 
\hline
\textbf{Our study}                      & \textbf{Yes (5)} & \textbf{Yes} & \textbf{Yes} & \textbf{Yes} & \textbf{Yes} & \textbf{Yes} \\
\hline
\end{tabular}
\end{table}

\textbf{Pretrained Foundation-Model Baselines for Visual Classification}: Zero-shot and lightweight vision foundation models eliminates or minimizes training by leveraging pretrained visual or vision-language representations. We categorize with three paradigms: 1) contrastive learning (CL), where image and text embeddings are aligned via a similarity objective; 2) knowledge-distillation self-supervised learning (KD-SSL), where a student network is trained to match a teacher's visual features without text supervision; and 3) autoregressive VLMs, which generate text conditioned on visual input. 

Radford et al.~\cite{radford2021learning} trained a joint image-text encoder on 400M collected pairs from the Internet and introduced CLIP.
Cherti et al.~\cite{cherti2023reproducible} reproduced the CLIP using ViT-L/14 on LAION-2B dataset and named it OpenCLIP.
Zhai et al.~\cite{zhai2023sigmoid} proposed SigLip: they have replaced the softmax contrastive loss with a per-pair sigmoid loss. This improved the effectiveness of the model's training and fine-grained recognition.
Oquab et al.~\cite{oquab2023dinov2} trained DINOv2 using self-supervised distillation, extending the self-supervised ViT approach of Caron et al.~\cite{caron2021dino}, but they didn't use any text supervision. This allowed them to generate good visual features, which opened up the possibility for the nearest-centroid classification.
Among autoregressive VLMs, Beyer et al.~\cite{beyer2024paligemma} combined a SigLIP vision encoder with the Gemma language model in PaliGemma (3B parameters).
Chu et al.~\cite{chu2024mobilevlm} designed MobileVLM~v2 for mobile deployment with a lightweight vision encoder and efficient language model.
Allal et al.~\cite{allal2025smolvlm} introduced SmolVLM, a compact VLM targeting resource-constrained settings.

\textbf{Multi-Objective Efficiency Metrics}: Schwartz et al.~\cite{schwartz2020green} presented their Green AI framework, that argued that efficiency should be as important a metric as accuracy.
Strubell et al.~\cite{strubell2019energy} measured the amount of energy consumed and CO$_2$ emissions when training NLP models. 
Patterson et al.~\cite{patterson2021carbon} expanded the previous analysis to large language and vision models, demonstrating the carbon footprint among different data centers.
Henderson et al.~\cite{henderson2020towards} proposed a system framework for informing the amount of energy and carbon dioxide as a separate metric in ML models and Thompson et al.~\cite{thompson2020computational} analyzed the trend of computational scaling and summed up that further growth is limited by the effectiveness of hardware.
Despite all this, there is no single standardized metric that combines accuracy, model complexity, memory, and inference cost on a common log-scale axis. 
To close this gap, we adopt the NetScore family of metrics (Section~\ref{sec:ns-metric}), extending the original formulation of Wong et al.~\cite{wong2018netscore,wong2019attonets,slnet2026} with two deployment-aware variants (NS$_{M}$, NS$^{\#}$) proposed in our paper.

Table~\ref{tab:related-work} compares this study with prior work across six dimensions. Our study addresses all three questions within a single framework: we compare five PEFT methods across Transformer and Mamba architectures, propose and evaluate an adaptive checkpointing algorithm, measure energy as a first-class metric, extend the NetScore family with two deployment-aware variants (NS$_{M}$, NS$^{\#}$) for energy- and memory-aware ranking, and benchmark zero-shot VLMs against fine-tuned models---all under an on-device VRAM budget (e.g., 2\,GB).
}


\begin{figure}[htbp]
\centering
\begin{tikzpicture}[
    >=Stealth,
    every node/.style={font=\footnotesize},
    block/.style={draw, rounded corners=2pt, fill=white,
                  text width=6.8cm, align=left, inner sep=3pt},
    paradigmTL/.style={fill=blue!6, rounded corners=3pt, draw=blue!35, thick},
    paradigmFM/.style={fill=orange!6, rounded corners=3pt, draw=orange!35, thick},
    bar/.style={draw, thick, rounded corners=3pt, minimum width=17cm,
                align=center, inner sep=4pt},
    result/.style={draw, rounded corners=3pt, fill=green!8, draw=green!50!black,
                   thick, text width=4.7cm, align=center, inner sep=3pt},
    flow/.style={->, thick, black!55, line width=0.7pt},
]

\node[bar, fill=gray!12] (data) at (0, 0) {%
    \textbf{Datasets}\;\textendash\;
    CIFAR-100 \textit{(100 cls, objects)}
    \quad$\big|$\quad
    DTD \textit{(47 cls, textures)}
};


\fill[blue!6, rounded corners=3pt] (-8.55, -0.85) rectangle (-0.25, -5.85);
\draw[paradigmTL] (-8.55, -0.85) rectangle (-0.25, -5.85);
\node[font=\footnotesize\bfseries, anchor=north] at (-4.4, -0.95)
    {Transfer Learning \;(Q1, Q2)};

\node[block] (peftarch) at (-4.4, -2.25) {%
    \textbf{PEFT $\times$ Model Arch.\ \;(Q1)}\\[1pt]
    \textit{PEFT:} Full FT $\cdot$ LoRA $\cdot$ AdaLoRA $\cdot$ QLoRA $\cdot$ BitFit\\[1pt]
    \textit{Backbones:} ViT-S, TinyViT (Transformer);\;
    Vim-S, MambaVision-T (Mamba)
};

\node[block] (ckpt) at (-4.4, -4.05) {%
    \textbf{Gradient Checkpointing \;(Q2)}\\[1pt]
    None $\cdot$ Static (per layer) $\cdot$ Adaptive (threshold $\tau$)
};

\node[draw, dashed, rounded corners=2pt, fill=yellow!10,
      text width=6.8cm, align=center, inner sep=3pt] (vram) at (-4.4, -5.4) {%
    \textit{Constraint:}\;on-device VRAM budget (e.g., 2\,GB)
};

\fill[orange!6, rounded corners=3pt] (0.25, -0.85) rectangle (8.55, -5.85);
\draw[paradigmFM] (0.25, -0.85) rectangle (8.55, -5.85);
\node[font=\footnotesize\bfseries, anchor=north] at (4.4, -0.95)
    {Foundation-Model Baselines \;(Q3)};

\node[block] (contrast) at (4.4, -2.15) {%
    \textbf{Contrastive VLMs}\\[1pt]
    OpenCLIP ViT-L/14 $\cdot$ SigLIP-Large
};

\node[block] (kdssl) at (4.4, -3.45) {%
    \textbf{KD-SSL Backbone}\\[1pt]
    DINOv2-Large \;(k-NN \,/\, linear probe)
};

\node[block] (ar) at (4.4, -4.95) {%
    \textbf{Autoregressive VLMs}\\[1pt]
    SmolVLM $\cdot$ MobileVLM\,v2 $\cdot$ PaliGemma-3B
};

\node[bar, fill=green!10] (metrics) at (0, -6.7) {%
    \textbf{Shared Metrics}\;\textendash\;
    Accuracy \;$\cdot$\; FT Time \;$\cdot$\; Energy (Wh) \;$\cdot$\;
    Peak VRAM \;$\cdot$\; NetScore family
};

\node[result] (R1) at (-5.7, -8.05) {%
    \textbf{Q1: Pareto frontier}\\[1pt]
    PEFT $\times$ Arch.\ trade-off\\[1pt]
    \textit{Tables~I, II, III}
};

\node[result] (R2) at (0, -8.05) {%
    \textbf{Q2: Memory--Energy}\\[1pt]
    Adaptive ckpt.\ trade-off\\[1pt]
    \textit{Tables~IV, V}
};

\node[result] (R3) at (5.7, -8.05) {%
    \textbf{Q3: Paradigm comparison}\\[1pt]
    Fine-tune vs.\ baseline\\[1pt]
    \textit{Tables~VI, VII}
};

\draw[flow] (-4.4, -0.30) -- (-4.4, -0.85);
\draw[flow] ( 4.4, -0.30) -- ( 4.4, -0.85);

\draw[flow] (-4.4, -5.85) -- (-4.4, -6.40);
\draw[flow] ( 4.4, -5.85) -- ( 4.4, -6.40);

\draw[flow] (-5.7, -7.00) -- (R1.north);
\draw[flow] ( 0.0, -7.00) -- (R2.north);
\draw[flow] ( 5.7, -7.00) -- (R3.north);

\end{tikzpicture}
\caption{Methodology overview. Two evaluation paradigms (\textbf{transfer learning} with PEFT and adaptive checkpointing; \textbf{foundation-model baselines} covering contrastive VLMs, a KD-SSL backbone, and autoregressive VLMs) are compared on the same datasets under an on-device VRAM budget (e.g., 2\,GB), through a shared set of metrics, and yield three question-specific outputs at the bottom: a PEFT$\times$architecture Pareto frontier (Q1, Tables~\ref{tab:peft-cifar}--\ref{tab:arch-comparison}), the memory--energy trade-off of adaptive checkpointing (Q2, Tables~\ref{tab:memory-reduction}--\ref{tab:ckpt-overhead}), and a fine-tune vs.\ training-free paradigm comparison (Q3, Tables~\ref{tab:contrastive_results}--\ref{tab:vlm_results}).}
\label{fig:methodology-overview}
\end{figure}
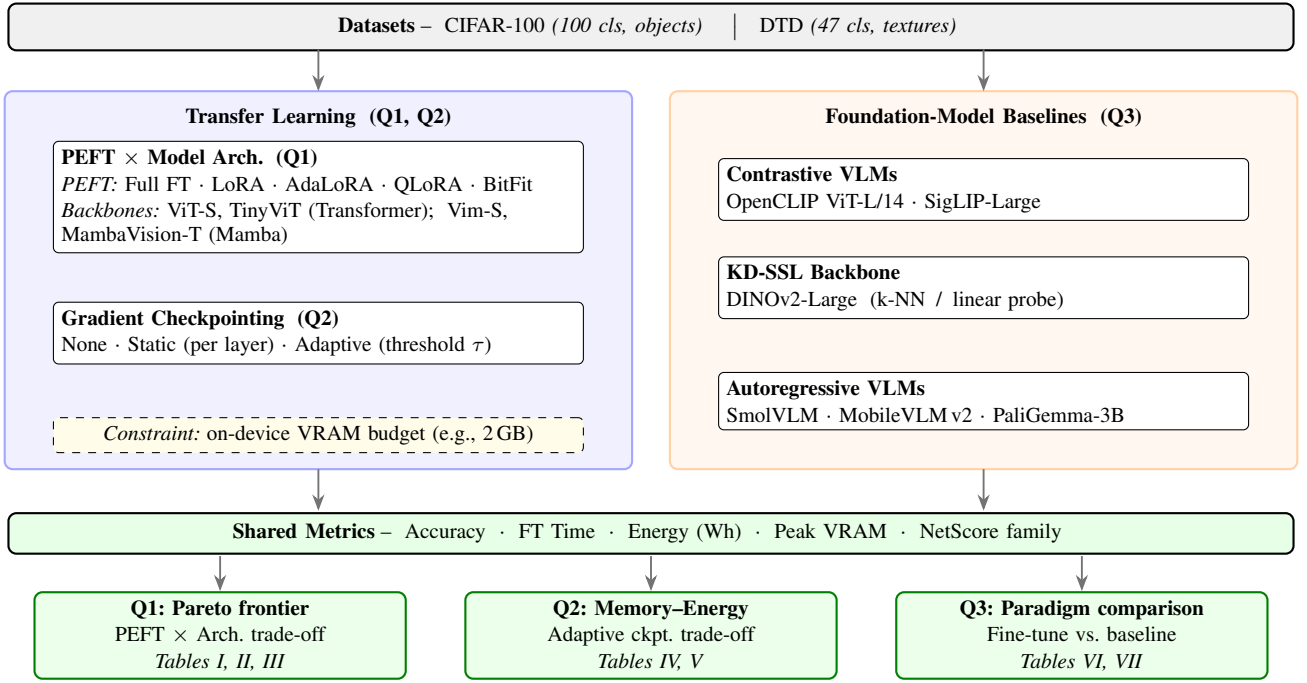


{
\chapter{Methodology}
\label{ch:methodology}

%



As shown in Figure~\ref{fig:methodology-overview}, it shows the general outline of our methodology. This study compares the two main paradigms: transfer learning and zero-shot learning. The experiments are conducted on identical datasets and use the same metrics. Each block corresponds to one or more questions.


{\textbf{Model Architectures $\times$ PEFT Methods (Q1)}: It is a joint design part over two axes that we vary together rather than independently.
The \emph{architecture} axis spans two families: Transformer-based backbones (ViT-Small, TinyViT) with quadratic attention complexity, and Mamba-based backbones (Vim-Small, MambaVision-Tiny) with linear-time sequence modeling --- chosen so that each family contributes one ``pure'' and one ``hybrid'' design and to keep the parameter budget within a comparable $21$--$32$\,M range.
The \emph{PEFT} axis spans five fine-tuning methods --- Full Fine-Tuning, LoRA ($r{=}8$, $\alpha{=}16$), AdaLoRA, QLoRA, and BitFit --- selected to span the trainable-parameter range from $<0.1\%$ (BitFit) to $100\%$ (Full FT) and to isolate distinct mechanisms (low-rank adaptation, adaptive rank allocation, 4-bit quantization, bias-only training).
Crossing the two axes yields $5 \times 4 = 20$ (method, architecture) configurations per dataset, with each cell evaluated on accuracy, FT time, energy, peak VRAM, and the NetScore family (Section~\ref{sec:ns-metric}). The joint structure is what allows Q1 to ask not just ``which PEFT method is best?'' or ``which architecture is best?'', but ``which \emph{combination} sits on the accuracy--energy Pareto frontier under a given on-device VRAM budget (e.g., 2\,GB)'', and the NetScore variants provide a unified multi-objective ranking across all 20 cells.}

\textbf{Gradient Checkpointing (Q2)}: Each transfer learning configuration was tested using the following 3 checkpoint modes: none, static (checkpointing is applied to each layer), and adaptive (using threshold). The adaptive checkpoint monitors GPU memory at runtime and can recalculate only those layers that exceeded the selected threshold. The main purpose of the adaptive checkpoint is to reduce the overhead of time and energy from a full static checkpoint, while maintaining the on-device VRAM budget (e.g., 2\,GB).

\textbf{Foundation-Model Baselines (Q3)}: Three families of pretrained models serve as training-free baselines: contrastive VLMs (SigLIP, OpenCLIP ViT-L/14), a KD-SSL model (DINOv2-large), and autoregressive VLMs (SmolVLM, PaliGemma-3B, MobileVLM~v2). By comparing their accuracy and inference energy against fine-tuned models, Q3 assesses whether task-specific PEFT fine-tuning is justified over zero-shot alternatives.

\textbf{Evaluation on Quantitative Metrics}: All configurations are compared using four metrics: top-1 test accuracy, training time, energy consumption (Wh), and the NetScore family (Section~\ref{sec:ns-metric}). 
}

{%
\section{Model Architectures (Q1)}
\label{sec:model-architectures}

The architectural axis of Q1 spans four backbones drawn from two paradigms: Transformer-based and Mamba-based, with one ``pure'' and one ``hybrid'' design per family. All four share a comparable parameter budget ($21$--$32$\,M) so that observed accuracy--energy differences reflect architectural mechanism rather than capacity.

\textbf{ViT-Small/16}~\cite{dosovitskiy2020image} ($22$\,M parameters) is the pure-Transformer baseline: $12$ encoder blocks, embedding dimension $384$, $6$ attention heads, $16{\times}16$ patch tokens, ImageNet-21k pretraining followed by ImageNet-1k fine-tuning. Attention is $\mathcal{O}(N^2)$ in the token count $N$, and the depth is shallow relative to the Mamba backbones, which together set its compute and memory footprint.

\textbf{TinyViT-21M}~\cite{wu2022tinyvit} ($21$\,M parameters) is a hybrid Transformer with an MBConv convolutional stem followed by windowed self-attention, pretrained on ImageNet-21k with knowledge distillation from a larger ViT. The MBConv stem allocates substantially larger spatial activations than a pure Transformer at the same parameter count, which becomes a load-bearing factor in the memory-reduction analysis (Section~\ref{sec:memory-reduction}).

\textbf{Vim-Small}~\cite{zhu2024vision} ($26$\,M parameters) is the pure-Mamba baseline: $24$ bidirectional Mamba blocks with linear-time sequence modeling, pretrained on ImageNet-1k. The sequential bidirectional pass and the deeper $24$-block stack determine its compute profile relative to the $12$-block Transformer baselines.

\textbf{MambaVision-Tiny}~\cite{hatamizadeh2024mambavision} ($32$\,M parameters) is a hybrid CNN$+$Mamba design with a convolutional stem and $12$ mixed Mamba/Transformer blocks, pretrained on ImageNet-1k. It pairs the linear-time Mamba mixer with a Transformer-style global path, occupying a different point in the Transformer--Mamba design space than Vim-Small.

\textbf{Selection rationale.}
This four-architecture set was chosen so that each paradigm contributes both a pure and a hybrid representative within a single parameter band, enabling within-family comparison (ViT-S vs.\ TinyViT; Vim-S vs.\ MambaVision-T) as well as across-paradigm comparison at matched capacity. A structural per-architecture description is sufficient here; the question of which architecture is most efficient under PEFT and a given on-device VRAM budget (e.g., 2\,GB) is answered jointly with the PEFT axis in Section~\ref{ch:results} (Q1).
}

\section{Gradient Checkpointing Methods (Q2)}
\label{sec:checkpointing-methods}



\textbf{Memory-budget selection.}
Despite the fact that the memory budget can be dynamically changed depending on the size of the VRAM, 
in this study we have decided to use the 2\,GB VRAM limit. The main reason is that this is directly related to two widely used on-device configurations: NVIDIA Jetson
Nano (4\,GB shared system/GPU memory), which is actively used in embedded systems and robotics, as well as an entry-level consumer GPU Nvidia GTX 1650(4\,GB in general, but the user usually has ${\sim}1.5$--$2$\,GB available due to the needs of operating systems, drivers and background processes).
We specifically chose the stricter 2\,GB limit because a configuration that is energy-efficient and feasible at 2\,GB will also be feasible at 4, 6, or 8\,GB. Crucially, the methodology is parameterised by the budget $M$, not a fixed number: the adaptive trigger $\tau\cdot M$ (Section~\ref{sec:adaptive-checkpointing}) and the peak-memory feasibility test both scale linearly with $M$, so the entire pipeline reproduces at any larger budget. We expect the relative ordering of PEFT methods and architectures to remain as $M$ grows; only feasibility changes (fewer OOMs, less need for checkpointing).

Training under the on-device VRAM budget (here 2\,GB) requires gradient checkpointing: TinyViT (21M parameters) exceeds the memory limit at batch size~32 without it. We compare two strategies---Static-1 and our proposed Adaptive method.

\subsection{Static-1 Checkpointing}
\label{sec:static-checkpointing}

Static-1 applies checkpointing to each layer. All intermediate activations are not saved during the forward pass and are recalculated again during backpropagation. This allows us to use a minimum amount of activation memory, but adds a maximum of recomputation overhead: exactly one additional forward pass per training step. 
If we assume that the standard cost of recalculating a model in which the backward pass costs ${\sim}2\times$ of the forward pass~\cite{chen2016training} (gradients with respect to both inputs and weights), then the baseline cost of training the model will be $F + B \approx F + 2F = 3F$ , where $F$ denotes the forward-pass FLOPs. Static-1 adds one full recomputation, which increases the step cost to $F + F_{\text{recomp}} + 2F = 4F$, a ${\sim}33\%$ step-time overhead which is equal to ${\sim}2\times$ the forward-pass FLOPs per step. Our measurements on ViT-S and TinyViT (Section~\ref{sec:timing-overhead}) show 17--44\% wall-clock energy overhead on both CIFAR-100 and DTD. This is consistent with this analytical bound once kernel-launch and memory-bandwidth effects are accounted for. Static-1 serves as a minimum boundary, because it is a standard approach in conditions of limited memory and does not require additional configuration.

{%
\subsection{Adaptive Checkpointing}
\label{sec:adaptive-checkpointing}

\paragraph{Algorithm.}
Our adaptive method decides \emph{per layer, per step} whether to checkpoint, based on the layer GPU memory usage. At each training step, it scans the forward pass in order; as soon as \texttt{torch.cuda.memory\_allocated()} exceeds a threshold $\tau\cdot M$ (where $M$ is the VRAM budget), a flag is raised and checkpointing is applied to the current layer and \emph{all subsequent layers} in that step. 
The flag resets at the next step, so the algorithm re-adapts to changing memory baselines (e.g., after optimizer-state allocation in step~1). 
Cascading rather than per-layer toggling is justified by a structural property of the forward pass: each layer stores its output activations in GPU memory until they are consumed by the corresponding backward-pass operation, so the cumulative activation memory is \emph{non-decreasing in the layer index}~$i$ during a single forward pass---it grows at layers that allocate new activations and stays flat at layers that do not (non-decreasing is the property we rely on; no linearity assumption is made, and the per-layer increment varies across blocks, 
once the threshold $\tau\cdot M$ is crossed at some layer $i^\star$ (i.e., $\texttt{mem\_allocated}()>\tau\cdot M$ for the first time), the non-decreasing property above guarantees that the same condition holds at every subsequent layer $i>i^\star$, so re-testing the condition past $i^\star$ would never disable checkpointing again; cascading from $i^\star$ to the end of the forward pass is therefore equivalent to per-layer toggling but avoids the redundant memory queries. Here $\tau\in(0,1]$ is a calibrated threshold ratio (see \emph{Calibration} below); intuitively, $\tau\cdot M$ is the level of allocated memory at which we decide to start trading compute for memory rather than risking an OOM later in the pass. 
Algorithm~\ref{alg:adaptive-ckpt} gives the full procedure. 

\begin{algorithm}[t]
\caption{Adaptive gradient checkpointing (one training step).}
\label{alg:adaptive-ckpt}
\begin{algorithmic}[1]
\Require layers $L = (l_1, \dots, l_N)$; budget $M$; threshold $\tau \in (0, 1]$
\State $\textit{ckpt} \gets \textsc{False}$
\For{$i = 1$ to $N$}
    \If{$\textit{ckpt}$ \textbf{is} \textsc{False} \textbf{and} $\texttt{mem\_allocated}() > \tau \cdot M$}
        \State $\textit{ckpt} \gets \textsc{True}$
    \EndIf
    \If{$\textit{ckpt}$}
        \State run $l_i$ under \texttt{torch.utils.checkpoint}
    \Else
        \State run $l_i$ normally (retain activations)
    \EndIf
\EndFor
\end{algorithmic}
\end{algorithm}

\paragraph{Role and Calibration of $\tau$.}
The $\tau$ threshold regulates how aggressively a checkpoint is used during training. 
A low $\tau$ causes a checkpoint on each layer, which entails recalculating many layers. Accordingly, the behavior tends to Static-1 as $\tau\to 0$. In this case, we are trading speed for memory safety.
However, a high $\tau$ delays the checkpoint until we have almost used up the entire memory budget, which is why a few (or none) layers are recalculated. That is, the behavior tends to no-checkpointing when $\tau\to 1$. This minimizes energy overhead, but it causes an OOM problem if the first "late layer" has already crossed ~$M$ (amplified by caching-allocator fragmentation) in memory. 
Accordingly, the important point is to find the \emph{largest} $\tau$ that still fits in ~$M$ with a safety margin.

Therefore, we are trying to calibrate $\tau$ empirically using descending linear sweep due to the fact that no prior work was reported as a principled value for live-memory-triggered checkpoint.
Starting from $\tau = 0.9$ (almost the minimum checkpoint) we run 5 training steps and accept $\tau$ if there is no OOM error. It is also important that peak memory is below $0.9\cdot M$, leaving a 10\% margin for caching-allocator fragmentation.
The 0.9\,$M$ margin is respected on ViT-S, MambaVision-T, and Vim-S; on TinyViT, only the LoRA/AdaLoRA/QLoRA variants satisfy it, while the heterogeneous MBConv stem on Full-FT and the freezing methods with large spatial activations (e.g., BitFit at 3\,057\,MiB) can breach the budget even at $\tau=0.5$ (Section~\ref{sec:memory-profiles}). 
If either condition fails, $\tau$ is decreased by $0.1$ and the test is repeated. The procedure terminates at the largest $\tau$ satisfying both conditions. 
Across all four architectures (ViT-Small/16, TinyViT-21M, MambaVision-Tiny, Vim-Small) under the chosen budget (here 2\,GB), the sweep converged to $\tau = 0.5$: higher values produced sporadic OOMs under fragmentation, while lower values were unnecessary. TinyViT Full-FT is an exception: adaptive checkpointing at $\tau = 0.5$ triggers OOM at the nominal batch size of 32 because the MBConv stem's first-layer activation already approaches the budget; the run is therefore performed at batch size~16 (peak 1{,}760\,MiB), cf.\ Figure~\ref{fig:waterfall-tinyvit}.
A finer-grained sweep (e.g., step~$0.05$) could yield small per-architecture refinements; we leave this for future work.

\paragraph{Scope of applicability.}
Adaptive checkpoint assumes a near-uniform per-layer activation footprint, so the cascade starts at one layer $i^\star$ whose position is stable across steps. This holds for pure Transformers (ViT-S) and pure Mamba models (Vim-S, MambaVision-T), where Section~\ref{sec:ckpt-analysis} confirms an energy advantage of adaptive over static. It breaks on heterogeneous hybrids (e.g., TinyViT, which combines MBConv$+$Attention blocks): the first MBConv stage alone allocates more memory than several transformer layers, so the adaptive algorithm either OOMs or consumes more power than static---which remains the recommended choice in this regime. A stage-aware threshold $\tau_s$ that adapts to each architectural stage is left as future work (Section~\ref{sec:limitations}).

}




}{

\section{Evaluation Protocol for Foundation-Model Baselines (Q3)}
\label{sec:zs-eval}

\paragraph{Terminology: source vs.\ target.}
We distinguish two paradigms by what happens to the parameters of the pretrained model on the target task:
\begin{itemize}
\item \emph{Transfer learning} (Q1, Q2). The \emph{source} task is the upstream supervised or self-supervised pretraining of the backbone (ImageNet-21k for ViT-S; ImageNet-1k for TinyViT, MambaVision-T, and Vim-S). The \emph{target} task is image classification on CIFAR-100 or DTD. The pretrained backbone is adapted to the target task by training a subset of its parameters with one of the five PEFT methods. The resulting fine-tuned model after that is being evaluated on the target test set. Inference therefore uses task-adapted weights.
\item \emph{Foundation-model baselines} (Q3). The \emph{source} side is the original pretraining of each baseline (large-scale image--text contrastive pretraining for OpenCLIP and SigLIP; self-distillation pretraining for DINOv2; vision-language instruction-style pretraining for SmolVLM, MobileVLM~V2, and PaliGemma). There is \emph{no} backbone adaptation step on CIFAR-100 or DTD --- the contrastive and autoregressive families perform genuine zero-shot inference, while the KD-SSL family (DINOv2) trains only a lightweight head ($k$-NN index or linear probe) on top of frozen features. ``Training-free'' here therefore means no backbone gradient updates and no target-task supervision of the representation, not zero exposure to natural images during pretraining.
\end{itemize}

\paragraph{Inference protocol per baseline family.}
Each baseline family produces a label using a different mechanism, but none of them updates the pretrained backbone on the target task.
\begin{itemize}
\item \emph{Contrastive VLMs} (OpenCLIP ViT-L/14, SigLIP-Large). Each candidate class name is wrapped in a fixed prompt template (``a photo of a \texttt{<class>}'') and encoded by the text tower; the test image is encoded by the vision tower; the predicted label is the class whose text embedding has the highest cosine similarity to the image embedding. This is genuine zero-shot inference.
\item \emph{KD-SSL} (DINOv2-Large). The frozen vision encoder produces a feature vector for every train and test image; classification is then performed via $k$-nearest-neighbor ($k$NN) and a linear probe on top of these features. The linear probe is the only trainable component, and its training is excluded from the energy budget reported as ``inference'' to keep the comparison conservative; feature extraction over the full dataset is, however, included.
\item \emph{Autoregressive VLMs} (SmolVLM-256M, MobileVLM~V2-1.7B, PaliGemma-3B). The model is given the image and a textual prompt requesting a class label; the predicted label is parsed from the generated text. This per-image text generation is what makes inference energy of these models substantially higher than the other baseline families.
\end{itemize}

\paragraph{Evaluation.}
Foundation-model baselines are evaluated on the same test sets as the transfer-learning methods (CIFAR-100 and DTD) using identical preprocessing pipelines.
Since no backbone fine-tuning occurs, the relevant metrics are: (1)~top-1 test accuracy, (2)~inference time over the full test set, and (3)~inference energy consumption.
Backbone training time and training energy are zero by definition. For NetScore computation on foundation-model baselines, we use the inference energy as the energy term, enabling a direct accuracy--inference-cost comparison with the fine-tuned models (Tables~\ref{tab:peft-cifar} and~\ref{tab:peft-dtd}, Inf.~Time / Inf.~E columns).


This comparison highlights the trade-off between the zero backbone-training cost of the foundation-model baselines and the potentially higher accuracy of fine-tuned models.

}
{
\section{Evaluation Metrics}
\label{sec:evaluation-protocol}


\subsection{Single-Objective Metrics}
\label{sec:single-obj-metrics}

Each configuration is evaluated on four independent metrics:

\begin{itemize}
\item \textbf{Top-1 test accuracy (\%):} Classification accuracy on the held-out test set after the final training epoch.
\item \textbf{Training time (min):} Wall-clock time for the full 10-epoch training run.
\item \textbf{Energy consumption (Wh):} Total GPU energy measured via \texttt{pynvml} (see Section~\ref{sec:power-measurement}).
\item \textbf{Peak VRAM (MB):} Maximum \texttt{torch.cuda.memory\_allocated()} observed during training. Determines whether a configuration fits within the chosen VRAM budget (here 2\,GB).
\end{itemize}

\subsection{Multi-Objective Metric: NetScore}
\label{sec:ns-metric}
\label{sec:sam-metric}

Single-objective metrics cannot directly identify the best accuracy--efficiency trade-off across configurations that differ in parameters, memory, and inference cost. We therefore adopt the \emph{NetScore} family~\cite{wong2018netscore,slnet2026}, a unified log-scale composite:

\begin{equation}
\mathrm{NS} \;=\; 20\cdot\log_{10}\!\left(\frac{a^{2}}{(pm)^{\beta}\,r^{\gamma}\,t^{\delta}\,P^{\lambda}}\right),
\label{eq:netscore}
\end{equation}
where $a$ is top-1 accuracy (\%), $p$ parameters (M), $m$ per-image FLOPs (G), $r$ peak inference VRAM (MiB), $t$ inference time (s), and $P$ average inference power (W); higher NS is better. The exponents $(\beta,\gamma,\delta,\lambda)$ select which efficiency axes are penalised. For PEFT Tables~\ref{tab:peft-cifar}--\ref{tab:peft-dtd}, $p$ counts \emph{trainable} parameters; for the foundation-model baseline Tables~\ref{tab:contrastive_results}--\ref{tab:vlm_results}, $p$ counts total model parameters.


Table~\ref{tab:ns-variants} defines all ten variants used in this work. Tables~\ref{tab:arch-comparison},~\ref{tab:contrastive_results},~\ref{tab:vlm_results} report only the four-variant $1/8$-weight deployment-only subset (NS, NS$_{E}$, NS$_{M}$, NS$^{\#}$).

\begin{table}[h]
\centering
\caption{NetScore variant definitions ($S=20$, $\alpha=2$). Exponents $(\beta,\gamma,\delta,\lambda)$ apply to (params$\,\times\,$FLOPs, peak VRAM, inf.\ time, inf.\ power). $\dagger$ denotes variants proposed in this work. Tables~\ref{tab:peft-cifar}--\ref{tab:peft-dtd} report all ten variants; Tables~\ref{tab:arch-comparison},~\ref{tab:contrastive_results},~\ref{tab:vlm_results} report only the $1/8$-weight deployment-only subset (NS, NS$_{E}$, NS$_{M}$, NS$^{\#}$).}
\label{tab:ns-variants}
\footnotesize
\begin{tabular}{@{}lccccll@{}}
\toprule
\textbf{Variant} & $\beta$ & $\gamma$ & $\delta$ & $\lambda$ & \textbf{Penalises} & \textbf{Source} \\
\midrule
NS   & 0.5   & 0     & 0     & 0     & Params $\times$ FLOPs                 & \cite{wong2018netscore,wong2019attonets} \\
NS$^{+}$   & 0.5   & 0.125 & 0.125 & 0     & + VRAM + inf.\ time                   & \cite{slnet2026} \\
NS$^{++}$ $(1/8)$  & 0.5   & 0.125 & 0.125 & 0.125 & + VRAM + time + power                 & This work \\
NS$^{++}$ $(1/4)$  & 0.5   & 0.25  & 0.25  & 0.25  & NS$^{++}$, doubled weight             & This work \\
\midrule
NS$_{E}$ $(1/8)$   & 0     & 0     & 0.125 & 0.125 & Inference energy (time $\times$ power) & \cite{gonzalez2024green,shi2024attention} \\
NS$_{E}$ $(1/4)$   & 0     & 0     & 0.25  & 0.25  & NS$_{E}$, doubled                     & \cite{gonzalez2024green,shi2024attention} \\
NS$_{M}$ $(1/8)$ $\dagger$  & 0     & 0.125 & 0     & 0     & Peak VRAM                             & This work \\
NS$_{M}$ $(1/4)$ $\dagger$  & 0     & 0.25  & 0     & 0     & NS$_{M}$, doubled                     & This work \\
NS$^{\#}$ $(1/8)$ $\dagger$ & 0     & 0.125 & 0.125 & 0.125 & VRAM + time + power                   & This work \\
NS$^{\#}$ $(1/4)$ $\dagger$ & 0     & 0.25  & 0.25  & 0.25  & NS$^{\#}$, doubled                    & This work \\
\bottomrule
\end{tabular}
\end{table}

The upper block (NS, NS$^{+}$, NS$^{++}$, $\beta>0$) rewards parameter-efficient methods even at equal inference cost; NS and NS$^{+}$ extend the original NetScore framework~\cite{wong2018netscore,wong2019attonets,slnet2026}. The lower block ($\beta=0$) focuses on deployment cost: NS$_{E}$ captures inference energy via the time\,$\times$\,power proxy, motivated by green continual-learning analyses~\cite{gonzalez2024green,shi2024attention}; NS$_{M}$ (proposed here) isolates peak memory pressure under on-device VRAM constraints; and NS$^{\#}$ (proposed here) penalises all three deployment axes jointly. For Tables~\ref{tab:arch-comparison}--\ref{tab:vlm_results}, only the $1/8$-weight deployment-only subset is reported, as the $1/4$ variants show monotonically amplified gaps without changing per-model leaders.
\emph{Future direction:} a natural 6-month extension is a \emph{lifecycle NetScore} variant that adds a training-energy term to the denominator, so that methods trading higher training cost for cheaper deployment (e.g.\ quantization-aware PEFT) are credited on a full lifecycle basis rather than on inference cost alone; the reported $(t,P)$ inference axes already supply the deployment half of such a metric.

\subsection{GPU Power Measurement}
\label{sec:power-measurement}

GPU power is sampled at training step boundaries using \texttt{pynvml}. At the start and end of each step, the instantaneous power draw is queried via \texttt{nvmlDeviceGetPowerUsage()}. Step energy is computed by trapezoidal approximation:
\begin{equation}
E_{\text{step}} = \frac{P_{\text{start}} + P_{\text{end}}}{2} \times \Delta t_{\text{step}},
\label{eq:energy-step}
\end{equation}
where $\Delta t_{\text{step}}$ is the step duration in hours. Total training energy is $E_{\text{total}} = \sum_{\text{steps}} E_{\text{step}}$ (Wh). At typical step durations of 0.1--2\,s, this approximation is sufficiently accurate for comparative analysis.

\subsection{Memory Measurement}
\label{sec:memory-measurement}

Peak and per-layer memory are tracked using \texttt{torch.cuda.memory\_allocated()} using forward pre-hooks and backward hooks that are integrated into the block of each model. The peak value across the full training run detects whether a configuration satisfies the chosen VRAM budget (here 2\,GB). We also take into account that the caching CUDA allocator may reserve more memory than is currently actively allocated. In addition, fragmentation can cause OOM even below the specified budget. Therefore, we have set a 10\% safety margin in our adaptive checkpoint threshold.
}

{
\chapter{Experimental Setup}
\label{sec:setup}


\section{Hardware and Software Configurations}
\label{sec:hardware}


Table~\ref{tab:hw-sw} summarizes the hardware and software stack used in all experiments.

\begin{narrowtable}[htbp]
\centering
\caption{Hardware and software configuration.
}
\label{tab:hw-sw}
\footnotesize
\begin{tabular}{@{}ll@{}}
\toprule
\textbf{Component} & \textbf{Specification} \\
\midrule
\multicolumn{2}{@{}l}{\textit{Hardware}} \\
GPU             & NVIDIA RTX 2060 SUPER \\
Total VRAM      & 8\,GB GDDR6 \\
VRAM budget     & 2\,GB (PEFT) / 8\,GB (VLM) \\
\midrule
\multicolumn{2}{@{}l}{\textit{Software}} \\
Framework       & PyTorch 2.x~\cite{paszke2019pytorch} \\
Model zoo       & \texttt{timm}~\cite{rw2019timm} \\
PEFT library    & HuggingFace PEFT~\cite{peft2023} \\
Quantization    & \texttt{bitsandbytes}~\cite{dettmers2022optimizers} \\
Mamba kernels   & \texttt{mamba\_ssm}~\cite{zhu2024vision} \\
Power telemetry & \texttt{pynvml} (NVML) \\
Optimizer       & AdamW~\cite{loshchilov2018decoupled} \\
\bottomrule
\end{tabular}
\end{narrowtable}

All experiments are conducted on a single GPU to ensure that energy and timing measurements are directly comparable across configurations. The VRAM budget is enforced via \texttt{torch.cuda.set\_per\_process\_memory\_fraction()} for PEFT fine-tuning (here set to 2\,GB, simulating edge devices such as the NVIDIA Jetson Nano; the same mechanism applies to any chosen $M$). Foundation-model baselines use the full 8\,GB since they require no backbone training.

\subsection{Reproducibility}
\label{sec:reproducibility}

A fixed random seed is set for Python, NumPy, PyTorch, and CUDA at the start of each run. Deterministic mode is enabled via \texttt{torch.use\_deterministic\_algorithms(True)} and \texttt{cudnn.benchmark = False}. All configurations are executed sequentially on the same hardware to eliminate variation from thermal throttling or background processes.

\begin{narrowtable}[htbp]
\centering
\caption{Dataset summary and preprocessing pipelines.
}
\label{tab:dataset-comparison}
\footnotesize
\begin{tabular}{@{}lcc@{}}
\toprule
\textbf{Property} & \textbf{CIFAR-100} & \textbf{DTD} \\
\midrule
Task           & Object classif.          & Texture recogn. \\
Classes        & 100                      & 47 \\
Train / Test   & 50{,}000 / 10{,}000     & 3{,}760 / 1{,}880 \\
Resolution     & $32{\times}32$           & Variable \\
Domain shift   & Moderate                 & Large \\
\midrule
Train resize   & $224{\times}224$         & RRC(224, 0.8--1.0) \\
Train flip     & HFlip($p{=}0.5$)        & HFlip($p{=}0.5$) \\
Test resize    & $224{\times}224$         & 256\,$\to$\,CC(224) \\
Normalize      & \multicolumn{2}{c}{ImageNet mean \& std} \\
\bottomrule
\end{tabular}
\end{narrowtable}

\section{Datasets}
\label{sec:datasets}

We used two main datasets: CIFAR-100 for object classification and DTD for texture recognition. Their characteristics (resolution, differences from ImageNet, and overfitting risk) are described in detail in Table ~\ref{tab:dataset-comparison}.

\subsection{CIFAR-100}
\label{sec:cifar100}
CIFAR-100~\cite{krizhevsky2009learning}: CIFAR-100 consists of 60,000 RGB images $32{\times}32$ and includes 100 classes that are divided into 50,000 training and 10,000 test images. All images are upsampled to $224{\times}224$ and normalized with ImageNet~\cite{deng2009imagenet} statistics.
\subsection{DTD (Describable Textures Dataset)}
\label{sec:dtd}
DTD (Describable Textures Dataset)~\cite{cimpoi2014describing}: DTD consists of 5640 images of different resolutions and 47 classes. An average of 120 images per class.
The two datasets complement each other well: configurations that work well on two datasets demonstrate robustness to domain shift
and data shortage. ImageNet-level experiments were excluded because measuring total energy with a 2\,GB memory limit would have made comparison impossible.

\section{Training Settings}
\label{sec:training-protocol}

\subsection{Optimizer and Hyperparameters}
\label{sec:optimizer}

All experiments use the AdamW optimizer~\cite{loshchilov2018decoupled} with a weight decay of 0.05 and default momentum parameters ($\beta_1 = 0.9$, $\beta_2 = 0.999$). The base learning rate is $5 \times 10^{-4}$ for all PEFT methods and $1 \times 10^{-4}$ for the full fine-tuning low-LR variant. No learning rate scheduler is applied; the learning rate remains constant throughout training. The use of a cosine or linear decay schedule would likely enhance the final accuracy results but a constant learning rate provides a straightforward experimental approach which allows researchers to identify performance changes that stem from model and PEFT method and checkpointing strategy rather than from schedule interactions.

\subsection{Training Duration}
\label{sec:epochs}

All experiments run for 10 epochs. For CIFAR-100 with batch size 64, this corresponds to $10 \times \lceil 50{,}000 / 64 \rceil = 7{,}820$ training steps; for DTD with batch size 64, $10 \times \lceil 3{,}760 / 64 \rceil = 590$ steps. The 10-epoch training schedule allows the model to learn adaptation while keeping the training time suitable for all testing requirements. We acknowledge that longer training may change the relative ordering of methods---particularly for BitFit, which adapts slowly due to its minimal parameter budget---and note this as a limitation in Section~\ref{sec:limitations}.

\subsection{Batch Size}
\label{sec:batch-size}

Memory limitations allow us to test configurations with batch sizes 16, 32, and 64.  Configurations of models, PEFT methods, and batch sizes that exceed 2 GB, even when using gradient checkpointing, will cause OOM. Accordingly, such configurations are marked as infeasible. The batch size is an important element because it determines the amount of memory that activations need. In addition, batch size affects training dynamics, gradient distribution, model performance, and energy consumption.

{%
\section{Architecture Comparison (Q1)}
\label{sec:architecture-comparison}

Section~\ref{sec:model-architectures} described what each backbone \emph{is}; here we connect those structural properties to the accuracy--energy behavior reported in Section~\ref{ch:results}, so that the Q1 results read as consequences of design choices rather than four unrelated data points. \textbf{ViT-Small} is the accuracy leader and the lowest-energy Transformer because its ImageNet-21k pretraining transfers with the smallest target--source delta, its short patch-token sequence keeps attention cheap, and its depth-12 activation graph fits the 2\,GB budget under every PEFT method---so the PEFT choice is energy- rather than feasibility-driven. \textbf{TinyViT} trails by $1$--$3$\,pp and its MBConv stem inflates spatial activations, pushing BitFit and Full-FT above budget so that only the LoRA/AdaLoRA/QLoRA cluster is feasible, and its energy is uniformly higher than ViT-S. \textbf{MambaVision-Tiny} (hybrid) is the lowest-energy backbone---a convolutional stem plus only 12 mixer blocks---at a $5$--$9$\,pp accuracy cost from ImageNet-1k-only pretraining. \textbf{Vim-Small} (pure Mamba) recovers most of ViT-S's accuracy but pays a ${\sim}3$--$4\times$ energy premium from its 24 sequential blocks and breaches the 2\,GB budget on every PEFT method, so it requires checkpointing (and bs\,=\,16 for Full-FT). These four design points---pretraining scale, stem type, block count, and sequence length---explain the accuracy--energy ordering quantified in Section~\ref{sec:peft-metrics}.
}
}



\chapter{Results \& Analysis}
\label{ch:results}






\section{PEFT Quantitative Metrics (Q1)}
\label{sec:peft-metrics}

{%

Tables~\ref{tab:peft-cifar} and~\ref{tab:peft-dtd} compare all five PEFT methods across four architectures under the no-checkpointing baseline, isolating method differences from checkpointing effects, on CIFAR-100 and DTD, respectively. 



\begin{table}[!t]
\centering
\caption{PEFT results on CIFAR-100 (no checkpointing). Training protocol: Section~\ref{sec:training-protocol}; NetScore variants: Section~\ref{sec:ns-metric}, Table~\ref{tab:ns-variants}. \textbf{Bold}: best per model. \colorbox{LightCyan}{Shaded}: best across all models. \textsuperscript{$\ast$}Vim-S Full-FT uses bs\,=\,16 (default bs\,=\,32 OOMs) and is excluded from best-per-model selection.
}
\label{tab:peft-cifar}
\scriptsize
\resizebox{\textwidth}{!}{%
\begin{tabular}{@{}llr>{}rrrrrrr>{}rr>{}r>{}r>{}r>{}r>{}r>{}r>{}r>{}r>{}r>{}r@{}}
\toprule
\textbf{Model} & \textbf{Method} & \textbf{\shortstack{Params\\(M)}} & {\textbf{\shortstack{FLOPs\\(G)}}} & \textbf{Acc} & \textbf{\shortstack{FT Time\\(min)}} & \textbf{\shortstack{Avg Pwr\\(W)}} & \textbf{\shortstack{Energy\\(Wh)}} & \textbf{\shortstack{VRAM\\(MiB)}} & \textbf{\shortstack{Inf. Time\\(s)}} & {\textbf{\shortstack{Inf. Pwr\\(W)}}} & \textbf{\shortstack{Inf. E\\(Wh)}} & {\textbf{NS}} & {\textbf{NS$^{+}$}} & {\textbf{\shortstack{NS$^{++}$\\$1/8$}}} & {\textbf{\shortstack{NS$^{++}$\\$1/4$}}} & {\textbf{\shortstack{NS$_{E}$\\$1/8$}}} & {\textbf{\shortstack{NS$_{E}$\\$1/4$}}} & {\textbf{\shortstack{NS$_{M}$\\$1/8$}}} & {\textbf{\shortstack{NS$_{M}$\\$1/4$}}} & {\textbf{\shortstack{NS$^{\#}$\\$1/8$}}} & {\textbf{\shortstack{NS$^{\#}$\\$1/4$}}} \\
\midrule
\multirow{5}{*}{ViT-S}
 & Full FT  & 21.70                              & 4.6 & \cellcolor{LightCyan}\textbf{0.897} & 34.1          & 146.4          & 83.19          & 1{,}084                           & 21.3          & 144.2                               & 0.853          & 58.12                               & 47.21                               & 41.81                               & 25.51                               & \cellcolor{LightCyan}\textbf{69.39} & \cellcolor{LightCyan}\textbf{60.67} & 70.52                               & 62.94                               & 61.81                               & 45.50                               \\
 & LoRA     & 0.27                               & 4.6 & 0.885                               & 32.4          & 146.5          & 79.20          & 921                               & 27.3          & 149.9                               & 1.137          & 76.94                               & 65.94                               & 60.50                               & 44.06                               & 68.85                               & 59.82                               & 70.47                               & 63.06                               & 61.44                               & 45.00                               \\
 & AdaLoRA  & 0.37                               & 4.6 & \cellcolor{LightCyan}\textbf{0.897} & 37.7          & 146.9          & 92.25          & 911                               & 28.7          & 150.3                               & 1.198          & 75.80                               & 64.76                               & 59.32                               & 42.83                               & 69.02                               & 59.94                               & 70.71                               & 63.31                               & 61.63                               & 45.14                               \\
 & QLoRA    & 0.27                               & 4.6 & 0.876                               & \textbf{25.4} & \textbf{141.1} & \textbf{59.84} & 831                               & \textbf{19.9} & \cellcolor{LightCyan}\textbf{137.3} & \textbf{0.759} & 76.76                               & 66.21                               & 60.87                               & 44.98                               & 69.11                               & 60.52                               & 70.40                               & 63.10                               & 61.81                               & 45.92                               \\
 & BitFit   & \cellcolor{LightCyan}\textbf{0.09} & 4.6 & 0.884                               & 26.2          & 146.2          & 63.85          & \cellcolor{LightCyan}\textbf{648} & 21.1          & 150.0                               & 0.879          & \cellcolor{LightCyan}\textbf{81.69} & \cellcolor{LightCyan}\textbf{71.35} & \cellcolor{LightCyan}\textbf{65.91} & \cellcolor{LightCyan}\textbf{50.13} & 69.11                               & 60.36                               & \cellcolor{LightCyan}\textbf{70.83} & \cellcolor{LightCyan}\textbf{63.80} & \cellcolor{LightCyan}\textbf{62.08} & \cellcolor{LightCyan}\textbf{46.30} \\
\midrule
\multirow{5}{*}{TinyViT}
 & Full FT  & 20.68 & 4.3 & \textbf{0.885} & 39.2          & 149.2                               & 108.65         & 3{,}787          & 27.3          & \textbf{146.8} & \textbf{1.113} & 58.39          & 45.85          & 40.43          & 22.48          & \textbf{68.87} & \textbf{59.86} & 68.93          & 59.99          & 59.92          & 41.97          \\
 & LoRA     & 0.24  & 4.3 & 0.871          & 31.1          & 142.7                               & 87.13          & 1{,}278          & 33.3          & 148.3          & 1.372          & 77.46          & 65.89          & 60.46          & 43.46          & 68.37          & 59.13          & 69.83          & 62.07          & 60.60          & 43.60          \\
 & AdaLoRA  & 0.34  & 4.3 & 0.872          & 35.5          & 142.4                               & 97.86          & 1{,}281          & 34.6          & 149.5          & 1.437          & 75.97          & 64.35          & 58.92          & 41.86          & 68.34          & 59.05          & 69.85          & 62.08          & 60.57          & 43.51          \\
 & QLoRA    & 0.24  & 4.3 & 0.871          & \textbf{28.3} & \cellcolor{LightCyan}\textbf{135.4} & \textbf{74.00} & \textbf{1{,}213} & 27.3          & 151.7          & 1.150          & 77.46          & 66.16          & 60.71          & 43.96          & 68.56          & 59.51          & \textbf{69.89} & \textbf{62.18} & \textbf{60.85} & \textbf{44.10} \\
 & BitFit   & 0.13  & 4.3 & 0.860          & 31.3          & 141.4                               & 84.43          & 3{,}307          & \textbf{27.2} & 149.2          & 1.127          & \textbf{79.91} & \textbf{67.52} & \textbf{62.09} & \textbf{44.27} & 68.36          & 59.34          & 68.58          & 59.78          & 59.56          & 41.74          \\
\midrule
\multirow{5}{*}{\shortstack[l]{Mamba-\\Vision-T}}
 & Full FT  & 31.79        & 4.4 & 0.804          & 31.0                               & 143.4          & 76.12                               & 2{,}223          & 20.6                               & \textbf{144.7} & 0.828                               & 54.75          & 43.10          & 37.70          & 20.65          & 67.52          & 58.84          & 67.84          & 59.48          & 59.16          & 42.10          \\
 & LoRA     & 1.13         & 4.4 & \textbf{0.839} & 25.1                               & 144.9          & 60.60                               & 1{,}336          & 23.9                               & 145.2          & 0.964                               & 69.99          & 58.72          & 53.32          & 36.65          & 68.10          & 59.25          & \textbf{69.14} & 61.32          & 60.29          & 43.62          \\
 & AdaLoRA  & 1.38         & 4.4 & 0.835          & 27.6                               & 145.1          & 66.74                               & 1{,}351          & 24.6                               & 146.9          & 1.004                               & 69.03          & 57.73          & 52.31          & 35.59          & 67.97          & 59.08          & 69.04          & 61.21          & 60.15          & 43.42          \\
 & QLoRA    & 1.13         & 4.4 & 0.835          & \cellcolor{LightCyan}\textbf{23.9} & \textbf{137.0} & \cellcolor{LightCyan}\textbf{54.56} & \textbf{1{,}253} & \cellcolor{LightCyan}\textbf{18.7} & 150.6          & \cellcolor{LightCyan}\textbf{0.782} & 69.90          & 58.98          & 53.53          & 37.16          & \textbf{68.24} & \textbf{59.62} & 69.12          & \textbf{61.38} & \textbf{60.50} & \textbf{44.13} \\
 & BitFit   & ${\sim}0.14$ & 4.4 & 0.837          & \textbf{24.2}                      & 143.7          & 58.01                               & 1{,}458          & 20.6                               & 146.1          & 0.836                               & \textbf{79.01} & \textbf{67.82} & \textbf{62.41} & \textbf{45.80} & 68.21          & 59.52          & 69.00          & 61.09          & 60.30          & 43.70          \\
\midrule
\multirow{5}{*}{Vim-S}
 & Full FT\textsuperscript{$\ast$} & 25.45 & 5.1 & 0.865 & 57.4 & 143.6 & 137.33 & 3{,}764 & 66.5 & 146.7 & 2.709 & 56.35 & 42.85 & 37.44 & 18.52 & 67.51 & 57.53 & 68.54 & 59.60 & 58.57 & 39.66 \\
 & LoRA     & 0.95         & 5.1 & 0.866          & 88.6          & 142.4          & 239.24          & 6{,}618          & 73.5          & 147.3          & 3.008          & 70.65          & 56.43          & 51.01          & 31.37          & 67.41          & 57.33          & 67.95          & 58.40          & 57.86          & 38.22          \\
 & AdaLoRA  & 1.39         & 5.1 & \textbf{0.872} & 90.8          & 147.0          & 252.01          & 6{,}619          & 71.3          & 154.0          & 3.050          & 69.11          & 54.93          & 49.46          & 29.81          & 67.52          & 57.42          & 68.07          & 58.52          & 57.97          & 38.31          \\
 & QLoRA    & 0.95         & 5.1 & 0.861          & 91.0          & 145.4          & 253.95          & 6{,}538          & 83.9          & \textbf{146.2} & 3.408          & 70.55          & 56.20          & 50.79          & 31.03          & 67.18          & 56.96          & 67.86          & 58.32          & 57.64          & 37.88          \\
 & BitFit   & ${\sim}0.08$ & 5.1 & 0.865          & \textbf{76.7} & \textbf{141.6} & \textbf{205.27} & \textbf{5{,}230} & \textbf{62.1} & 149.6          & \textbf{2.580} & \textbf{81.37} & \textbf{67.60} & \textbf{62.16} & \textbf{42.94} & \textbf{67.56} & \textbf{57.64} & \textbf{68.18} & \textbf{58.89} & \textbf{58.26} & \textbf{39.05} \\
\bottomrule
\end{tabular}
}
\end{table}

\subsection{Results on CIFAR-100} As shown in Table~\ref{tab:peft-cifar}, PEFT results on CIFAR-100 are summarized.

\textbf{Accuracy}: ViT-S achieves the highest accuracy across all methods (0.897), followed by TinyViT (0.885), Vim-S (0.872), and MambaVision-T (0.839). 
Vim-S Full-FT exceeds the 2\,GB budget without checkpointing (OOM), but PEFT methods complete successfully at 5.1--6.5\,GB.

{%
\emph{Protocol and batch-size clarification.} All rows in Tables~\ref{tab:peft-cifar}--\ref{tab:peft-dtd} use bs\,=\,32 except the Vim-S Full-FT row (marked~\textsuperscript{$\ast$}), which halves the batch to bs\,=\,16 because Vim-S Full-FT OOMs at the 2\,GB budget at bs\,=\,32. At bs\,=\,16 the peak VRAM drops to 3{,}764\,MiB (still above 2\,GB) and training converges, confirming the OOM is a memory-budget rather than an architectural effect; the row is excluded from best-per-model marking because the smaller batch shifts the gradient-noise regime and is not directly comparable to the other entries.
The 2\,GB budget was chosen over 4\,GB precisely to create discriminating pressure on Full-FT: TinyViT Full-FT peaks at 3{,}787\,MiB and Vim-S Full-FT at 3{,}764\,MiB (bs\,=\,16), both of which fit at 4\,GB and would collapse the Full-FT vs.\ PEFT vs.\ checkpointing trade-off into the no-checkpointing column. The target is motivated by on-device hardware (Jetson Nano with 4\,GB \emph{shared} system/GPU memory; GTX~1650 with ${\sim}1.5$--$2$\,GB usable for a workload after OS/driver overhead; see Section~\ref{sec:checkpointing-methods}); rankings obtained at 2\,GB generalize upward since the pipeline is parameterized by the budget $M$.
}

The accuracy gap between the best PEFT method and Full-FT depends on the architecture. On ViT-S, the difference is not so strong (AdaLoRA correlates with Full-FT by 0.897), but on MambaVision-T PEFT, the methods \emph{outperform} Full-FT by 0.035 (LoRA 0.839 vs. Full-FT 0.804), where freezing is the largest backbone (32M parameters) gives a stronger regularization. 


The two Mamba models show different behavior: MambaVision-T(hybrid, 32M parameters and 12 Mamba/Transformer blocks) shows that PEFT methods outperform Full-FT by 0.035 on CIFAR-100 (LoRA 0.839 vs.\ Full-FT 0.804).  This is significantly more than on ViT-S or TinyViT, which shows stronger regulation from freezing a larger model. However, the absolute accuracy of the MambaVision-T is 0.058 lower than that of the ViT-S on the CIFAR-100. (On DTD, the difference is not so significant - 0.006(0.772 vs. 0.778)).



{%
\textbf{FT-Time}:
Fine-tuning time on CIFAR-100 (Fig.~\ref{fig:fttime-cifar}) is dominated by the architecture, not the PEFT method: the three Transformer/hybrid backbones cluster in a $24$--$39$\,min band while Vim-S sits $2.5$--$3\times$ higher ($76.7$--$91.0$\,min). Within each backbone QLoRA/BitFit sit near the minimum and Full-FT/AdaLoRA near the maximum, mirroring the energy ranking; across all 20 configurations the range is $73.7\%$ ($23.9$--$91.0$\,min, a $3.81\times$ gap).

\begin{narrowfigure}[!t]
\centering
\begin{tikzpicture}
\begin{axis}[
    name=fttimeplot,
    width=0.78\linewidth,
    height=4.3cm,
    scale only axis,
    xlabel={PEFT method},
    ylabel={FT Time (min)},
    symbolic x coords={Full FT,LoRA,AdaLoRA,QLoRA,BitFit},
    xtick=data,
    ymin=20, ymax=95,
    axis y line*=left,
    axis x line*=bottom,
    grid=major,
    grid style={dashed,gray!30},
    legend style={at={(0.5,-0.32)},anchor=north,legend columns=4,
                  font=\scriptsize,/tikz/every even column/.append style={column sep=0.25cm},
                  draw=gray!50,fill opacity=0.9,draw opacity=1,text opacity=1,inner sep=2pt},
    tick label style={font=\scriptsize},
    label style={font=\footnotesize},
    mark size=1.5pt,
]
\addplot[blue,mark=*,thick] coordinates {(Full FT,34.1)(LoRA,32.4)(AdaLoRA,37.7)(QLoRA,25.4)(BitFit,26.2)};
\addlegendentry{ViT-S}
\addplot[red,mark=square*,thick] coordinates {(Full FT,39.2)(LoRA,31.1)(AdaLoRA,35.5)(QLoRA,28.3)(BitFit,31.3)};
\addlegendentry{TinyViT}
\addplot[teal,mark=triangle*,thick] coordinates {(Full FT,31.0)(LoRA,25.1)(AdaLoRA,27.6)(QLoRA,23.9)(BitFit,24.2)};
\addlegendentry{MambaVision-T}
\addplot[orange,mark=diamond*,thick] coordinates {(Full FT,57.4)(LoRA,88.6)(AdaLoRA,90.8)(QLoRA,91.0)(BitFit,76.7)};
\addlegendentry{Vim-S}
\end{axis}
\begin{axis}[
    at={(fttimeplot.south west)},
    anchor=south west,
    width=0.78\linewidth,
    height=4.3cm,
    scale only axis,
    symbolic x coords={Full FT,LoRA,AdaLoRA,QLoRA,BitFit},
    xtick=data,
    ymin=0.78, ymax=0.92,
    axis y line*=right,
    axis x line=none,
    ylabel={},
    tick label style={font=\scriptsize},
    label style={font=\footnotesize},
    mark size=1.3pt,
]
\addplot[blue,mark=o,dashed,thick,forget plot] coordinates {(Full FT,0.897)(LoRA,0.885)(AdaLoRA,0.897)(QLoRA,0.876)(BitFit,0.884)};
\addplot[red,mark=square,dashed,thick,forget plot] coordinates {(Full FT,0.885)(LoRA,0.871)(AdaLoRA,0.872)(QLoRA,0.871)(BitFit,0.860)};
\addplot[teal,mark=triangle,dashed,thick,forget plot] coordinates {(Full FT,0.804)(LoRA,0.839)(AdaLoRA,0.835)(QLoRA,0.835)(BitFit,0.837)};
\addplot[orange,mark=diamond,dashed,thick,forget plot] coordinates {(Full FT,0.865)(LoRA,0.866)(AdaLoRA,0.872)(QLoRA,0.861)(BitFit,0.865)};
\end{axis}
\end{tikzpicture}
\caption{Fine-tuning time (solid, left axis) and accuracy (dashed, right axis) on CIFAR-100 across PEFT methods (x-axis: Full~FT $\to$ LoRA $\to$ AdaLoRA $\to$ QLoRA $\to$ BitFit; method-family grouping). Colors match the legend; dashed line for each architecture overlays its CIFAR-100 accuracy. AdaLoRA carries $\approx 1.37\!\times$ the parameters of LoRA (see Tables~\ref{tab:peft-cifar},~\ref{tab:peft-dtd}); QLoRA matches LoRA exactly. Architecture dominates the time spread: Transformer/hybrid backbones cluster in $24$--$39$\,min, Vim-S is $2.5$--$3\!\times$ slower. Accuracy is far flatter: ViT-S/TinyViT span $\le 2.5$\,pp across PEFT methods, MambaVision-T spans $3.5$\,pp, Vim-S only $1.1$\,pp. $^{\ast}$Vim-S Full~FT uses bs\,=\,16 (bs\,=\,32 OOMs at the 2\,GB budget) and is not directly comparable.}
\label{fig:fttime-cifar}
\end{narrowfigure}

\textbf{Time--accuracy joint view (e.g., MambaVision-T)}:
The dashed accuracy curves overlaid in Fig.~\ref{fig:fttime-cifar} reveal that the time--accuracy trade-off is not uniform across architectures.
MambaVision-T is the clearest case: Full-FT requires the longest training of any MV-T configuration ($31.0$\,min) yet yields the \emph{lowest} CIFAR-100 accuracy ($0.804$), whereas LoRA/AdaLoRA/QLoRA/BitFit reach $0.835$--$0.839$ in $23.9$--$27.6$\,min --- PEFT Pareto-dominates Full-FT on this backbone (shorter training \emph{and} higher accuracy simultaneously).
ViT-S and TinyViT show a milder pattern, with Full-FT marginally best in accuracy ($0.5$--$2.5$\,pp gap) but $\sim$5--8\,min slower than the cheapest PEFT method.
Vim-S accuracy is essentially flat ($\pm 0.006$ around $0.866$) despite a $\sim$15\,min training-time spread, so PEFT-vs-Full-FT is decided by time/energy rather than accuracy.
The practical implication is that the dense baseline pays its largest accuracy dividend on the two pure Transformers, while on the hybrid Mamba (MambaVision-T) and pure Mamba (Vim-S) PEFT methods are not just cheaper but also at least as accurate as Full-FT.


\textbf{Power Consumption}:
Average GPU power (Table~\ref{tab:peft-cifar}, Avg~Pwr column) is far more stable than fine-tuning time: per-model averages cluster within $142.2$--$145.4$\,W and the spread across PEFT methods never exceeds $9.2\%$ ($135.4$--$149.2$\,W over all 20 configurations). Power thus behaves as an architecture-level constant ($\pm5$\,W), so the per-method energy differences in Table~\ref{tab:peft-cifar} track the FT-time differences above almost one-to-one rather than any power signature.}

\textbf{Accuracy--Energy Trade-Offs}:
On every backbone the Pareto-optimal configurations are QLoRA and BitFit without checkpointing: they stay within $0.01$--$0.02$ accuracy of Full-FT while consuming $20$--$30\%$ less energy (e.g., ViT-S QLoRA $59.84$\,Wh, $28\%$ below Full-FT, at $0.876$ vs.\ $0.897$). The two Mamba backbones occupy opposite corners: MambaVision-T minimizes energy at an accuracy cost, whereas Vim-S recovers accuracy---its PEFT methods reach $0.861$--$0.872$ versus $0.804$--$0.839$ for MambaVision-T---but is markedly energy-intensive, with Vim-S BitFit drawing $205$\,Wh against $64$\,Wh for ViT-S BitFit ($3.2\times$) and $58$\,Wh for MambaVision-T BitFit ($3.5\times$). The gap is driven by training time ($2.9\times$ longer than ViT-S at a similar ${\sim}142$\,W), reflecting the cost of 24 sequential Mamba blocks versus 12 Transformer layers; Vim-S Full-FT additionally OOMs without checkpointing.



{%
\textbf{Trainable parameters vs.\ peak VRAM}:
A common assumption in the PEFT literature is that fewer trainable parameters automatically translate into lower peak VRAM. Fig.~\ref{fig:vram-cifar} shows that this correlation is non-monotonic and architecture-dependent.
ViT-S follows the assumption: peak VRAM decreases monotonically from Full-FT through LoRA/AdaLoRA/QLoRA to BitFit, all comfortably under the $2$\,GB budget.
TinyViT breaks the ordering: BitFit has the \emph{fewest} trainable parameters but the \emph{second-highest} peak ($\sim\!2.6\!\times$ above the LoRA/AdaLoRA/QLoRA cluster), and only the LoRA/AdaLoRA/QLoRA group fits the budget.
Vim-S inverts the assumption sharply: BitFit sits well above the budget while LoRA/AdaLoRA/QLoRA peak even higher despite carrying more trainable parameters; the only Vim-S configuration that approaches the $4$\,GB regime is Full-FT$^{\ast}$ at bs\,=\,16, despite having the largest trainable count.

The reason is that, at bs\,=\,32 and $224\!\times\!224$ input, peak VRAM is dominated by the activation graph rather than by the parameter or optimizer state. PEFT methods that route gradients through a few localized adapter modules (LoRA/AdaLoRA/QLoRA) allow the autograd engine to discard activations from layers without trainable parameters; BitFit, by contrast, trains a bias term in \emph{every} layer, which keeps the activation graph alive across the full network and inflates the peak. QLoRA further removes weight memory through 4-bit quantization, but on Vim-S and TinyViT this saving is small relative to the activation footprint.
The practical takeaway is that for memory planning under a tight VRAM budget, the right predictor is \emph{where} the trainable parameters live (which layers carry gradients), not how many of them there are.

\begin{narrowfigure}[!t]
\centering
\begin{tikzpicture}
\begin{axis}[
    name=vramplot,
    width=0.78\linewidth,
    height=4.3cm,
    scale only axis,
    xlabel={PEFT method},
    ylabel={Peak VRAM (MiB)},
    symbolic x coords={Full FT,LoRA,AdaLoRA,QLoRA,BitFit},
    xtick=data,
    ymin=0, ymax=7200,
    axis y line*=left,
    axis x line*=bottom,
    grid=major,
    grid style={dashed,gray!30},
    legend style={at={(0.5,-0.32)},anchor=north,legend columns=4,
                  font=\scriptsize,/tikz/every even column/.append style={column sep=0.25cm},
                  draw=gray!50,fill opacity=0.9,draw opacity=1,text opacity=1,inner sep=2pt},
    tick label style={font=\scriptsize},
    label style={font=\footnotesize},
    mark size=1.5pt,
]
\addplot[blue,mark=*,thick] coordinates {(Full FT,1084)(LoRA,921)(AdaLoRA,911)(QLoRA,831)(BitFit,648)};
\addlegendentry{ViT-S}
\addplot[red,mark=square*,thick] coordinates {(Full FT,3787)(LoRA,1278)(AdaLoRA,1281)(QLoRA,1213)(BitFit,3307)};
\addlegendentry{TinyViT}
\addplot[teal,mark=triangle*,thick] coordinates {(Full FT,2223)(LoRA,1336)(AdaLoRA,1351)(QLoRA,1253)(BitFit,1458)};
\addlegendentry{MambaVision-T}
\addplot[orange,mark=diamond*,thick] coordinates {(Full FT,3764)(LoRA,6618)(AdaLoRA,6619)(QLoRA,6538)(BitFit,5230)};
\addlegendentry{Vim-S}
\draw[red,dashed,thick] ({rel axis cs:0,0}|-{axis cs:Full FT,2048}) -- ({rel axis cs:1,0}|-{axis cs:Full FT,2048});
\node[red,font=\scriptsize,anchor=south east] at ({rel axis cs:1,0}|-{axis cs:Full FT,2048}) {2\,GB budget};
\end{axis}
\begin{axis}[
    at={(vramplot.south west)},
    anchor=south west,
    width=0.78\linewidth,
    height=4.3cm,
    scale only axis,
    symbolic x coords={Full FT,LoRA,AdaLoRA,QLoRA,BitFit},
    xtick=data,
    ymin=0.78, ymax=0.92,
    axis y line*=right,
    axis x line=none,
    ylabel={},
    tick label style={font=\scriptsize},
    label style={font=\footnotesize},
    mark size=1.3pt,
]
\addplot[blue,mark=o,dashed,thick,forget plot] coordinates {(Full FT,0.897)(LoRA,0.885)(AdaLoRA,0.897)(QLoRA,0.876)(BitFit,0.884)};
\addplot[red,mark=square,dashed,thick,forget plot] coordinates {(Full FT,0.885)(LoRA,0.871)(AdaLoRA,0.872)(QLoRA,0.871)(BitFit,0.860)};
\addplot[teal,mark=triangle,dashed,thick,forget plot] coordinates {(Full FT,0.804)(LoRA,0.839)(AdaLoRA,0.835)(QLoRA,0.835)(BitFit,0.837)};
\addplot[orange,mark=diamond,dashed,thick,forget plot] coordinates {(Full FT,0.865)(LoRA,0.866)(AdaLoRA,0.872)(QLoRA,0.861)(BitFit,0.865)};
\end{axis}
\end{tikzpicture}
\caption{Peak VRAM (solid, left axis) and accuracy (dashed, right axis) on CIFAR-100 across PEFT methods (x-axis: Full~FT $\to$ LoRA $\to$ AdaLoRA $\to$ QLoRA $\to$ BitFit; method-family grouping; AdaLoRA carries $\approx 1.37\!\times$ the parameters of LoRA, see Tables~\ref{tab:peft-cifar},~\ref{tab:peft-dtd}). Colors match the legend. Red dashed line: $2$\,GB budget. The VRAM-vs-accuracy relationship is non-monotonic: ViT-S follows the conventional ordering (VRAM decreases from Full~FT to BitFit, accuracy nearly flat at $0.876$--$0.897$), TinyViT inverts it for BitFit (lowest accuracy at high VRAM), and Vim-S inverts it sharply (LoRA/AdaLoRA/QLoRA exceed BitFit in VRAM yet accuracy is flat). Average VRAM is reported for ViT-S/TinyViT and peak VRAM for MambaVision-T/Vim-S, mirroring Table~\ref{tab:peft-cifar}. $^{\ast}$Vim-S Full~FT uses bs\,=\,16 and is not directly comparable to the other entries.}
\label{fig:vram-cifar}
\end{narrowfigure}

\textbf{VRAM--accuracy joint view (e.g., MambaVision-T)}:
The dashed accuracy curves overlaid in Fig.~\ref{fig:vram-cifar} make the memory--accuracy trade-off explicit.
On ViT-S the cheapest configuration in memory (BitFit, $648$\,MiB) reaches $0.884$ accuracy --- only $1.3$\,pp below Full-FT ($1{,}084$\,MiB, $0.897$) --- so accuracy and VRAM are largely decoupled.
TinyViT shows an unfavourable corner: BitFit pays the second-highest VRAM cost ($3{,}307$\,MiB, breaching the $2$\,GB budget) and simultaneously delivers the lowest accuracy ($0.860$); the LoRA/AdaLoRA/QLoRA cluster at $\sim$$1{,}260$\,MiB and $0.871$--$0.872$ Pareto-dominates both BitFit and Full-FT.
MambaVision-T is the most striking case: \emph{every} PEFT configuration ($1{,}253$--$1{,}458$\,MiB at $0.835$--$0.839$) Pareto-dominates Full-FT ($2{,}223$\,MiB at $0.804$) --- lower VRAM \emph{and} higher accuracy.
Vim-S decouples the two axes entirely: accuracy is essentially flat ($0.861$--$0.872$) across a $\sim$$2.4$\,GB VRAM spread, none of the configurations fits the $2$\,GB budget, and the choice is driven purely by memory.
The general lesson is that the conventional VRAM ordering predicts accuracy on ViT-S only; on the other three backbones, accuracy is decided by where trainable parameters live, and the Pareto frontier moves toward PEFT.

}

{\textbf{NetScore Analysis (CIFAR-100)}:
Across the ten NetScore variants (Section~\ref{sec:ns-metric}, Table~\ref{tab:ns-variants}), the complexity-aware \textbf{NS} column rewards BitFit on every backbone (cross-all leader $81.69$) because it attains near-best accuracy at the smallest parameter$\times$FLOPs footprint, whereas the inference-energy \textbf{NS$_E$} column is accuracy-driven: it tips to Full-FT on ViT-S ($69.39$) but to QLoRA on MambaVision-T ($68.24$), where PEFT also out-accuracies Full-FT. The VRAM-penalized \textbf{NS$_M$}/\textbf{NS$^{\#}$} variants favour the smallest-footprint method per backbone and demote the TinyViT BitFit anomaly ($3{,}307$\,MiB). Because all tables use single-pass inference energy, NS$_E$ also ranks across paradigms: ViT-S Full-FT leads at $69.39$, ahead of OpenCLIP ($67.35$) and DINOv2 Linear ($65.73$), since fine-tuned small backbones pay only $0.85$\,Wh per pass versus $3.19$--$35.50$\,Wh for the foundation-model baselines---so once the one-time FT cost is paid, they dominate inference efficiency whenever inference repeats.}

{%
\textbf{Key Findings PEFT on CIFAR-100}:
The results point to several architecture-aware deployment guidelines rather than a single dominant method. When VRAM and energy budgets are tight, low-rank or bias-only adapters (QLoRA, BitFit) should be preferred over full fine-tuning: they occupy the Pareto frontier of the accuracy--energy trade-off on every architecture studied, with only a marginal accuracy cost on Transformers and, for hybrid Mamba backbones, with an accuracy \emph{advantage} attributable to the regularization effect of freezing a larger pretrained backbone. Conversely, full fine-tuning becomes attractive only when the backbone is a small, well-matched pretrained Transformer and energy is not a concern.

A consistent architecture-dependent pattern tells you when to use Full-FT instead of PEFT. On pure Transformers, the PEFT--Full-FT difference is not significant, so Full-FT does not offer any practical benefit for its increased power consumption. On hybrid Mamba--Transformer models, PEFT is rather the recommended standard. Due to the freezing of the model, energy decreases and regularization improves. On a pure Mamba model with deep sequential block, Full-FT is infeasible under a tight on-device budget (here 2\,GB), and PEFT methods are energy-intensive. Accordingly, Mamba models should be selected only when accuracy is justified for more significant energy consumption.

{The NetScore variants (Section~\ref{sec:ns-metric}) capture this two-regime structure explicitly. NS (complexity-aware, $\beta\!=\!0.5$) always rewards the adapter with the smallest parameter$\!\times\!$FLOPs footprint---BitFit on every backbone---and is the appropriate variant under tight deployment-footprint constraints. NS$_E$ (inference-energy-aware, $\beta\!=\!0$) surfaces accuracy differences and tips to Full-FT or the accuracy-leading PEFT method on each backbone, making it the right choice when a 1--2 accuracy-point premium is worth the marginal inference-time increase; the gap between NS and NS$_E$ leaders is narrow enough (${\le}0.28$ on ViT-S) that the ranking is sensitive to the deployment priority. The rapid convergence of pre-trained Transformers on coarse-grained datasets further suggests that epoch budgets can be significantly reduced with an early-stopping criterion, which expands the advantage of PEFT under both variants.}}

\begin{table}[!t]
\centering
\caption{PEFT results on DTD (no checkpointing); columns, units, and conventions identical to Table~\ref{tab:peft-cifar}. Training protocol: Section~\ref{sec:training-protocol}; NetScore variants: Section~\ref{sec:ns-metric}, Table~\ref{tab:ns-variants}. \textbf{Bold}: best per model. \colorbox{LightCyan}{Shaded}: best across all models. \textsuperscript{$\ast$}Vim-S Full-FT uses bs\,=\,16 and is excluded from best-per-model selection.
}
\label{tab:peft-dtd}
\scriptsize
\resizebox{\textwidth}{!}{%
\begin{tabular}{@{}llr>{}rrrrrrr>{}rr>{}r>{}r>{}r>{}r>{}r>{}r>{}r>{}r>{}r>{}r@{}}
\toprule
\textbf{Model} & \textbf{Method} & \textbf{\shortstack{Params\\(M)}} & {\textbf{\shortstack{FLOPs\\(G)}}} & \textbf{Acc} & \textbf{\shortstack{FT Time\\(min)}} & \textbf{\shortstack{Avg Pwr\\(W)}} & \textbf{\shortstack{Energy\\(Wh)}} & \textbf{\shortstack{VRAM\\(MiB)}} & \textbf{\shortstack{Inf. Time\\(s)}} & {\textbf{\shortstack{Inf. Pwr\\(W)}}} & \textbf{\shortstack{Inf. E\\(Wh)}} & {\textbf{NS}} & {\textbf{NS$^{+}$}} & {\textbf{\shortstack{NS$^{++}$\\$1/8$}}} & {\textbf{\shortstack{NS$^{++}$\\$1/4$}}} & {\textbf{\shortstack{NS$_{E}$\\$1/8$}}} & {\textbf{\shortstack{NS$_{E}$\\$1/4$}}} & {\textbf{\shortstack{NS$_{M}$\\$1/8$}}} & {\textbf{\shortstack{NS$_{M}$\\$1/4$}}} & {\textbf{\shortstack{NS$^{\#}$\\$1/8$}}} & {\textbf{\shortstack{NS$^{\#}$\\$1/4$}}} \\
\midrule
\multirow{5}{*}{ViT-S}
 & Full FT  & 21.68                              & 4.6 & 0.764                               & 29.9          & 143.9          & 71.63          & 1{,}128                           & 3.9          & \textbf{144.0} & 0.156          & 55.34                               & 46.23                               & 40.83                               & 26.33                               & 68.45                               & 61.58                               & 67.69                               & 60.06                               & 60.82                               & 46.32                               \\
 & LoRA     & 0.25                               & 4.6 & 0.774                               & 29.2          & 140.5          & 68.45          & 946                               & 5.3          & 144.7          & 0.213          & 74.94                               & 65.69                               & 60.29                               & 45.64                               & 68.34                               & 61.13                               & 68.11                               & 60.67                               & 60.90                               & 46.25                               \\
 & AdaLoRA  & 0.35                               & 4.6 & \cellcolor{LightCyan}\textbf{0.778} & 34.3          & 140.3          & 80.29          & 969                               & 5.5          & 145.3          & 0.222          & 73.57                               & 64.25                               & 58.85                               & 44.13                               & 68.38                               & 61.13                               & 68.17                               & 60.71                               & 60.92                               & 46.19                               \\
 & QLoRA    & 0.25                               & 4.6 & 0.766                               & \textbf{22.9} & 141.1          & \textbf{53.91} & 878                               & \textbf{3.8} & 146.8          & \textbf{0.155} & 74.76                               & 65.95                               & 60.54                               & 46.31                               & \cellcolor{LightCyan}\textbf{68.50} & \textbf{61.64}                      & 68.01                               & 60.65                               & 61.14                               & 46.92                               \\
 & BitFit   & \cellcolor{LightCyan}\textbf{0.07} & 4.6 & 0.767                               & 23.8          & 141.8          & 56.20          & \cellcolor{LightCyan}\textbf{687} & 4.0          & 146.7          & 0.163          & \cellcolor{LightCyan}\textbf{80.31} & \cellcolor{LightCyan}\textbf{71.72} & \cellcolor{LightCyan}\textbf{66.30} & \cellcolor{LightCyan}\textbf{52.29} & 68.47                               & 61.55                               & \cellcolor{LightCyan}\textbf{68.30} & \cellcolor{LightCyan}\textbf{61.21} & \cellcolor{LightCyan}\textbf{61.38} & \cellcolor{LightCyan}\textbf{47.36} \\
\midrule
\multirow{5}{*}{TinyViT}
 & Full FT  & 20.65 & 4.3 & \textbf{0.777} & 40.0          & 143.1                               & 97.48          & 4{,}018          & 5.2          & \textbf{144.7} & \textbf{0.209} & 56.13          & 45.33          & 39.93          & 23.73          & \textbf{68.43} & \textbf{61.23} & 66.61          & 57.60          & 59.42          & 43.21          \\
 & LoRA     & 0.21  & 4.3 & 0.772          & 31.6          & 137.6                               & 74.88          & 1{,}339          & 6.3          & 144.6          & 0.253          & 75.95          & 66.13          & 60.73          & 45.52          & 68.11          & 60.71          & 67.69          & 59.87          & 60.29          & 45.07          \\
 & AdaLoRA  & 0.30  & 4.3 & 0.775          & 35.3          & 144.7                               & 87.81          & 1{,}344          & 6.6          & 147.8          & 0.271          & 74.47          & 64.60          & 59.17          & 43.88          & 68.10          & 60.63          & \textbf{67.75} & 59.93          & 60.28          & 44.98          \\
 & QLoRA    & 0.21  & 4.3 & 0.769          & \textbf{28.7} & \cellcolor{LightCyan}\textbf{133.0} & \textbf{65.52} & \textbf{1{,}258} & \textbf{5.2} & 149.5          & 0.216          & 75.88          & 66.34          & 60.90          & 45.93          & 68.21          & 60.98          & 67.69          & \textbf{59.94} & \textbf{60.46} & \textbf{45.49} \\
 & BitFit   & 0.10  & 4.3 & 0.760          & 31.8          & 136.5                               & 74.23          & 3{,}513          & 5.2          & 146.8          & 0.212          & \textbf{78.90} & \textbf{68.24} & \textbf{62.83} & \textbf{46.76} & 68.03          & 60.82          & 66.37          & 57.50          & 59.16          & 43.09          \\
\midrule
\multirow{5}{*}{\shortstack[l]{Mamba-\\Vision-T}}
 & Full FT  & 31.79        & 4.4 & 0.757          & 28.2                               & 144.1          & 67.88                               & 2{,}231          & 3.7                               & 147.9                               & 0.152                               & 53.71          & 43.91          & 38.49          & 23.27          & 68.32          & 61.47                               & 66.79          & 58.42          & 59.95          & 44.73          \\
 & LoRA     & 1.13         & 4.4 & \textbf{0.772} & 21.4                               & 154.2          & 54.49                               & 1{,}337          & 4.3                               & 154.0                               & 0.184                               & 68.54          & 59.14          & 53.67          & 38.80          & 68.45          & 61.40                               & \textbf{67.69} & \textbf{59.87} & \textbf{60.64} & 45.77          \\
 & AdaLoRA  & 1.37         & 4.4 & 0.771          & 23.7                               & 154.7          & 61.05                               & 1{,}352          & 4.4                               & 160.4                               & 0.196                               & 67.68          & 58.24          & 52.73          & 37.78          & 68.36          & 61.24                               & 67.65          & 59.83          & 60.53          & 45.58          \\
 & QLoRA    & 1.13         & 4.4 & 0.752          & 21.0                               & 144.7          & \cellcolor{LightCyan}\textbf{50.50} & \textbf{1{,}251} & \cellcolor{LightCyan}\textbf{3.4} & \cellcolor{LightCyan}\textbf{137.7} & \cellcolor{LightCyan}\textbf{0.130} & 68.08          & 59.01          & 53.66          & 39.25          & 68.37          & \cellcolor{LightCyan}\textbf{61.70} & 67.31          & 59.56          & 60.63          & \textbf{46.21} \\
 & BitFit   & ${\sim}0.13$ & 4.4 & 0.761          & \cellcolor{LightCyan}\textbf{20.9} & 150.4          & 52.40                               & 1{,}452          & 3.7                               & 141.1                               & 0.145                               & \textbf{77.68} & \textbf{68.36} & \textbf{62.98} & \textbf{48.28} & \textbf{68.46} & 61.67                               & 67.35          & 59.45          & 60.56          & 45.86          \\
\midrule
\multirow{5}{*}{Vim-S}
 & Full FT\textsuperscript{$\ast$} & 25.43 & 5.1 & 0.748 & 47.8 & 141.9 & 113.04 & 3{,}764 & 12.3 & 150.1 & 0.513 & 53.83 & 42.16 & 36.72 & 19.62 & 66.79 & 58.62 & 66.02 & 57.08 & 57.85 & 40.75 \\
 & LoRA     & 0.95         & 5.1 & 0.718          & 92.0          & 139.1          & 218.74          & 6{,}618          & 15.8          & 148.8          & 0.653          & 67.39          & 54.84          & 49.41          & 31.43          & 65.82          & 57.39          & 64.69          & 55.14          & 56.27          & 38.29          \\
 & AdaLoRA  & 1.37         & 5.1 & \textbf{0.755} & 91.9          & 141.1          & 221.57          & 6{,}618          & 13.6          & 148.5          & 0.561          & 66.67          & 54.29          & 48.86          & 31.05          & \textbf{66.85} & \textbf{58.59} & \textbf{65.57} & \textbf{56.01} & \textbf{57.30} & 39.49          \\
 & QLoRA    & 0.95         & 5.1 & 0.736          & 90.1          & 144.2          & 223.09          & 6{,}537          & 15.5          & \textbf{144.2} & 0.621          & 67.82          & 55.31          & 49.91          & 32.00          & 66.30          & 57.93          & 65.14          & 55.60          & 56.76          & 38.85          \\
 & BitFit   & ${\sim}0.07$ & 5.1 & 0.721          & \textbf{78.0} & \textbf{139.2} & \textbf{185.64} & \textbf{5{,}230} & \textbf{11.7} & 146.8          & \textbf{0.477} & \textbf{78.79} & \textbf{66.82} & \textbf{61.41} & \textbf{44.02} & 66.23          & 58.14          & 65.02          & 55.72          & 56.93          & \textbf{39.55} \\
\bottomrule
\end{tabular}
}
\end{table}

{%
\subsection{Results on DTD}


\textbf{Accuracy}: As shown in Table~\ref{tab:peft-dtd}, the best-case accuracies across the four models are: ViT-S (0.778, AdaLoRA), TinyViT (0.777, Full-FT), MambaVision-T (0.772, LoRA), and Vim-S (0.755, AdaLoRA). The spread between the best and worst PEFT configurations is 0.060 (0.718 on Vim-S LoRA vs.\ 0.778 on ViT-S AdaLoRA), substantially narrower than the 0.093 spread on CIFAR-100, reflecting the fact that DTD's small training set (${\sim}$3{,}760 images) limits how much architectural differences can manifest. Vim-S Full-FT again OOMs without checkpointing.

The architecture-dependent PEFT--Full-FT gap is more pronounced and, in contrast to CIFAR-100, PEFT dominates on three of four architectures. On ViT-S, AdaLoRA \emph{exceeds} Full-FT by 0.014 (0.778 vs.\ 0.764). On MambaVision-T, LoRA exceeds Full-FT by 0.015 (0.772 vs.\ 0.757), mirroring the CIFAR-100 regularization effect on the 32M-parameter backbone. On Vim-S, AdaLoRA exceeds the best LoRA/QLoRA/BitFit by 0.019--0.037. TinyViT is the only exception, where Full-FT (0.777) narrowly outperforms the best PEFT method (AdaLoRA 0.775). The overall picture is that freezing the backbone serves as implicit regularization under the data-scarce DTD regime.

Both Mamba models show contrasting behavior on CIFAR-100. MambaVision-T (hybrid) compares almost exactly with ViT-S on DTD(0.006 vs.\ 0.058 on CIFAR-100), which is an assumption that hybrid Mamba--Transformer architectures are better suited for fine-grained textures than for coarse-grained object categories. At the same time, Vim-S (pure Mamba) loses to other architectures by 0.017--0.023, reversing its advantage over MambaVision-T, which was observed on CIFAR-100. This fact indicates that deeper sequential Mamba stacks cannot effectively transfer their pre-trained features to fine-grained domains with a limited amount of marked up data.

\textbf{Accuracy--Energy Trade-Offs}:
As on CIFAR-100, QLoRA and BitFit are the most energy-efficient: on ViT-S, QLoRA consumes $53.91$\,Wh ($25\%$ below Full-FT) while slightly exceeding its accuracy ($0.766$ vs.\ $0.764$), and TinyViT QLoRA saves $33\%$ ($65.52$\,Wh). TinyViT is again less efficient than the similarly-sized ViT-S (Full-FT $97.48$ vs.\ $71.63$\,Wh) because its MBConv stem inflates activation memory. Vim-S is the most energy-intensive backbone by a wide margin---BitFit draws $185.64$\,Wh versus $56.20$\,Wh for ViT-S BitFit ($3.3\times$), driven by a $3.3\times$ longer training time at near-identical power. The Pareto frontier is therefore anchored by QLoRA/BitFit on Transformers and LoRA/QLoRA on MambaVision-T; AdaLoRA is competitive only when accuracy dominates ($0.778$ on ViT-S, at $49\%$ higher energy than QLoRA), and no Vim-S configuration reaches it.

{%
\textbf{Time and memory on DTD}:
Fine-tuning time and peak VRAM follow the same architecture-dominated, non-monotonic pattern as CIFAR-100 (Figs.~\ref{fig:fttime-cifar},~\ref{fig:vram-cifar}). Transformer/hybrid backbones train in $20.9$--$40.0$\,min and Vim-S in $78.0$--$92.0$\,min, and the accuracy overlay is again PEFT-favored on three of four backbones: MambaVision-T LoRA reaches $0.772$ in $21.4$\,min and Pareto-dominates Full-FT ($0.757$ at $28.2$\,min), while ViT-S QLoRA gives up only $1.2$\,pp for a $33\%$ time saving. Peak VRAM is likewise activation-dominated: ViT-S stays under the $2$\,GB budget for all methods ($687$--$1{,}128$\,MiB), TinyViT BitFit again breaches it ($3{,}513$\,MiB despite the fewest trainable parameters), and every Vim-S PEFT configuration exceeds it ($5{,}230$--$6{,}618$\,MiB). The ``where, not how many'' memory lesson from CIFAR-100 transfers unchanged.
}

{\textbf{NetScore Analysis (DTD)}:
The NS ranking mirrors CIFAR-100 (BitFit wins every backbone; cross-all $80.31$), but NS$_E$ diverges informatively: on the data-scarce DTD, QLoRA overtakes Full-FT on ViT-S ($68.50$ vs.\ $68.45$) and leads cross-all because it is both marginally more accurate ($0.766$ vs.\ $0.764$) and the cheapest in energy ($53.91$\,Wh); at the $1/4$ weight MambaVision-T QLoRA leads across backbones ($61.70$). The lesson is that NS$_E$ is dataset-sensitive---a small accuracy edge on a scarce benchmark is magnified into a clear NS$_E$ win, whereas a larger benchmark can tip NS$_E$ toward Full-FT.}

\textbf{Training Dynamics}: On DTD the small training set (${\sim}$59 batches per epoch) makes per-epoch accuracy oscillate by $0.03$--$0.04$ (versus $<0.005$ on CIFAR-100 after epoch~5), and a run can even regress in the final epochs (e.g., ViT-S QLoRA drops from $0.764$ at epoch~5 to $0.737$ at epoch~10); this is small-dataset stochastic instability at a fixed learning rate, not a checkpoint-specific effect.

\textbf{Key Findings PEFT on DTD}:
The fine-grained DTD mode with data deficiency well supports the guidelines received at CIFAR-100. Further training on small datasets is much more noisy, and our results reason against choosing Full-FT by default. On the DTD dataset, PEFT methods are superior to Full-FT on all architectures. The reason for this is the freezing of a qualitatively pre-trained backbone, which is an implicit regularization mechanism in conditions where there is insufficient labeled data. The practical consequence is that when choosing a method on small datasets, priority should be given to PEFT first, leaving Full-FT for cases where the preconfigured representations do not match well with the target domain.

The same Pareto structure as on CIFAR-100 holds: QLoRA and BitFit deliver the best accuracy--energy trade-off on Transformers, while on hybrid Mamba backbones low-rank adapters (LoRA, AdaLoRA) are preferable when accuracy is the dominant criterion. Pure-Mamba backbones remain the least attractive choice under a tight on-device budget (here 2\,GB)---they incur a multiplicative energy penalty over Transformers without a compensating accuracy advantage on fine-grained data. {As on CIFAR-100, NS (complexity-aware) uniformly favours BitFit while NS$_E$ surfaces the energy--accuracy interaction specific to the dataset; on DTD the NS$_E$ winner is QLoRA rather than Full-FT, so the appropriate variant should be chosen based on whether deployment footprint (NS) or inference efficiency (NS$_E$) is the primary criterion.}

A regime-specific finding is that training dynamics on DTD are unstable: per-epoch accuracy oscillates noticeably and a single run can collapse in the final epoch despite strong earlier performance. The practical guideline is to adopt best-epoch checkpoint selection or validation-driven early stopping when labeled data are scarce, rather than relying on final-epoch metrics.

}

{\textbf{NetScore variant scope from this point onward.} Tables~\ref{tab:peft-cifar}--\ref{tab:peft-dtd} above report the full ten-variant NetScore family because the PEFT$\times$architecture grid is the only setting where all five efficiency axes (params, FLOPs, VRAM, time, power) vary independently across rows. Outside this grid the complexity-aware NS$^{+}$ and NS$^{++}$ degenerate: the architecture comparison (Table~\ref{tab:arch-comparison}) holds the trainable-parameter footprint fixed per PEFT method so the params$\times$FLOPs penalty is near-constant within each method row, and the foundation-model baseline tables (Tables~\ref{tab:contrastive_results}--\ref{tab:vlm_results}) have no PEFT axis at all, so NS, NS$^{+}$, and NS$^{++}$ produce identical per-table rankings; we therefore retain only NS. For the same simplicity reason we report only the $1/8$-weight versions of NS$_{E}$, NS$_{M}$, and NS$^{\#}$: empirically the $1/4$ versions monotonically amplify the same per-model ordering in log-space without rotating any leader (verified across all $20$ PEFT cells in Tables~\ref{tab:peft-cifar}--\ref{tab:peft-dtd}), and the softer $1/8$ weighting keeps accuracy as the dominant term, consistent with the original NetScore formulation~\cite{wong2018netscore,wong2019attonets}. The remainder of this chapter therefore reports only NS, NS$_{E}$, NS$_{M}$, and NS$^{\#}$ at the $1/8$ weight (cf.\ Table~\ref{tab:ns-variants}, Section~\ref{sec:ns-metric}).}

{%
\subsection{Model Comparison on CIFAR-100 and DTD}
\label{sec:arch-comparison}






Table~\ref{tab:arch-comparison} provides a direct comparison across all four models on both CIFAR-100 and DTD, using the no-checkpointing configuration to isolate architectural differences from checkpointing effects.

\begin{narrowtable}[htbp]
\centering
\caption{Cross-architecture comparison on CIFAR-100 and DTD (no checkpointing); data drawn from Tables~\ref{tab:peft-cifar}--\ref{tab:peft-dtd}. The four rightmost columns are the $1/8$-weight deployment-only NetScore subset (Section~\ref{sec:ns-metric}, Table~\ref{tab:ns-variants}). \textbf{Bold}: best per column within each (dataset, method); \underline{underline}: worst. Vim-S Full-FT OOMs and is excluded from ranking.
}
\label{tab:arch-comparison}
\footnotesize
\resizebox{\columnwidth}{!}{%
\begin{tabular}{@{}llrrr>{}r>{}r>{}r>{}r@{}}
\toprule
\textbf{Method} & \textbf{Model} & \textbf{Acc} & \textbf{\shortstack{Energy\\(Wh)}} & \textbf{\shortstack{VRAM\\(MiB)}} & {\textbf{NS}} & {\textbf{\shortstack{NS$_{E}$\\$1/8$}}} & {\textbf{\shortstack{NS$_{M}$\\$1/8$}}} & {\textbf{\shortstack{NS$^{\#}$\\$1/8$}}} \\
\midrule
\multicolumn{9}{c}{\textbf{CIFAR-100}} \\
\midrule
\multirow{4}{*}{Full-FT}
 & ViT-S        & \textbf{0.897} & 83.19  & \textbf{1{,}084} & 58.12 & \textbf{69.39} & \textbf{70.52} & \textbf{61.81} \\
 & TinyViT      & 0.885 & \underline{108.65} & \underline{3{,}787} & \textbf{58.39} & 68.87 & 68.93 & 59.92 \\
 & MambaVision-T & \underline{0.804} & \textbf{76.12}  & 2{,}223 & \underline{54.75} & \underline{67.52} & \underline{67.84} & \underline{59.16} \\
 & Vim-S        & \multicolumn{7}{c}{OOM} \\
\midrule
\multirow{4}{*}{LoRA}
 & ViT-S        & \textbf{0.885} & 79.20  & \textbf{921} & 76.94 & \textbf{68.85} & \textbf{70.47} & \textbf{61.44} \\
 & TinyViT      & 0.871 & 87.13  & 1{,}278 & \textbf{77.46} & 68.37 & 69.83 & 60.60 \\
 & MambaVision-T & \underline{0.839} & \textbf{60.60}  & 1{,}336 & \underline{69.99} & 68.10 & 69.14 & 60.29 \\
 & Vim-S        & 0.866 & \underline{239.24} & \underline{6{,}618} & 70.65 & \underline{67.41} & \underline{67.95} & \underline{57.86} \\
\midrule
\multirow{4}{*}{QLoRA}
 & ViT-S        & \textbf{0.876} & 59.84  & \textbf{831} & 76.76 & \textbf{69.11} & \textbf{70.40} & \textbf{61.81} \\
 & TinyViT      & 0.871 & 74.00  & 1{,}213 & \textbf{77.46} & 68.56 & 69.89 & 60.85 \\
 & MambaVision-T & \underline{0.835} & \textbf{54.56}  & 1{,}253 & \underline{69.90} & 68.24 & 69.12 & 60.50 \\
 & Vim-S        & 0.861 & \underline{253.95} & \underline{6{,}538} & 70.55 & \underline{67.18} & \underline{67.86} & \underline{57.64} \\
\midrule
\multirow{4}{*}{BitFit}
 & ViT-S        & \textbf{0.884} & 63.85  & \textbf{648} & \textbf{81.69} & \textbf{69.11} & \textbf{70.83} & \textbf{62.08} \\
 & TinyViT      & 0.860 & 84.43  & 3{,}307 & 79.91 & 68.36 & 68.58 & 59.56 \\
 & MambaVision-T & \underline{0.837} & \textbf{58.01}  & 1{,}458 & \underline{79.01} & 68.21 & 69.00 & 60.30 \\
 & Vim-S        & 0.865 & \underline{205.27} & \underline{5{,}230} & 81.37 & \underline{67.56} & \underline{68.18} & \underline{58.26} \\
\midrule
\multicolumn{9}{c}{\textbf{DTD}} \\
\midrule
\multirow{4}{*}{Full-FT}
 & ViT-S        & 0.764          & 71.63          & \textbf{1{,}128} & 55.34 & \textbf{68.45} & \textbf{67.69} & \textbf{60.82} \\
 & TinyViT      & \textbf{0.777} & \underline{97.48}  & \underline{4{,}018} & \textbf{56.13} & 68.43 & \underline{66.61} & \underline{59.42} \\
 & MambaVision-T & \underline{0.757} & \textbf{67.88} & 2{,}231          & \underline{53.71} & \underline{68.32} & 66.79 & 59.95 \\
 & Vim-S        & \multicolumn{7}{c}{OOM} \\
\midrule
\multirow{4}{*}{LoRA}
 & ViT-S        & \textbf{0.774} & 68.45          & \textbf{946}     & 74.94 & 68.34 & \textbf{68.11} & \textbf{60.90} \\
 & TinyViT      & 0.772          & 74.88          & 1{,}339          & \textbf{75.95} & 68.11 & 67.69 & 60.29 \\
 & MambaVision-T & 0.772         & \textbf{54.49} & 1{,}337          & 68.54 & \textbf{68.45} & 67.69 & 60.64 \\
 & Vim-S        & \underline{0.718} & \underline{218.74} & \underline{6{,}618} & \underline{67.39} & \underline{65.82} & \underline{64.69} & \underline{56.27} \\
\midrule
\multirow{4}{*}{QLoRA}
 & ViT-S        & 0.766          & 53.91          & \textbf{878}     & 74.76 & \textbf{68.50} & \textbf{68.01} & \textbf{61.14} \\
 & TinyViT      & \textbf{0.769} & 65.52          & 1{,}258          & \textbf{75.88} & 68.21 & 67.69 & 60.46 \\
 & MambaVision-T & 0.752         & \textbf{50.50} & 1{,}251          & 68.08 & 68.37 & 67.31 & 60.63 \\
 & Vim-S        & \underline{0.736} & \underline{223.09} & \underline{6{,}537} & \underline{67.82} & \underline{66.30} & \underline{65.14} & \underline{56.76} \\
\midrule
\multirow{4}{*}{BitFit}
 & ViT-S        & \textbf{0.767} & 56.20          & \textbf{687}     & \textbf{80.31} & \textbf{68.47} & \textbf{68.30} & \textbf{61.38} \\
 & TinyViT      & 0.760          & 74.23          & 3{,}513          & 78.90 & 68.03 & 66.37 & 59.16 \\
 & MambaVision-T & 0.761         & \textbf{52.40} & 1{,}452          & \underline{77.68} & 68.46 & 67.35 & 60.56 \\
 & Vim-S        & \underline{0.721} & \underline{185.64} & \underline{5{,}230} & 78.79 & \underline{66.23} & \underline{65.02} & \underline{56.93} \\
\bottomrule
\end{tabular}%
}
\end{narrowtable}

The four architectures clearly show a hierarchy in terms of accuracy and efficiency. ViT-S shows the highest accuracy among all PEFT methods (0.876--0.897) for the average energy cost (60--92\,Wh). Vim-S is second in accuracy (0.861--0.872) but last in energy efficiency (205--254\,Wh, 3--4$\times$ higher than ViT-S), which is why it shows the lowest deployment-aware NetScores (NS$_{E}$, NS$_{M}$, NS$^{\#}$) in every method group. The MambaVision-T offers the lowest amount of power consumption (55--76\,Wh), but significantly lower accuracy (0.804--0.839). TinyViT showed the worst results: lower accuracy than the ViT-S, higher energy than the MambaVision-t, the highest memory consumption among all Full-FT models (3{,}787\,MiB).

Two Mamba models have shown that ``Mamba'' is not always effective. The hybrid design of MambaVision-T (MBConv stem + 12 mixed blocks, 32M params) provides lower energy cost, but weaker pretrained features. On the other hand, Vim-S's pure bidirectional Mamba design (24 blocks, 26M params) preserves representations well, which is why accuracy approaches to the result of ViT-S. But due to the sequential processing of 24 blocks, the energy consumption increases significantly.

On DTD, the accuracy gap narrows: MambaVision-T LoRA reaches 0.772 versus ViT-S AdaLoRA at 0.778, while consuming 32\% less energy (54.49\,Wh vs 80.29\,Wh). Vim-S AdaLoRA achieves 0.755 on DTD---trailing both ViT-S and MambaVision-T---at 221.57\,Wh. This suggests that Vim-S's advantage over MambaVision-T is specific to larger datasets (CIFAR-100) where its richer representations matter more.

\textbf{Key Findings on Architectures}:
ViT-S dominates accuracy, VRAM, and nearly every deployment-aware NetScore column (NS$_{E}$, NS$_{M}$, NS$^{\#}$) across all PEFT methods, while the complexity-aware NS favors TinyViT's slightly smaller adapters in the LoRA/QLoRA groups; MambaVision-T offers the lowest energy but the lowest accuracy (and takes NS$_{E}$ on DTD in the LoRA/QLoRA/BitFit groups thanks to its cheapest inference).
The two Mamba models are not interchangeable: MambaVision-T's hybrid design trades accuracy for energy efficiency, while Vim-S's pure bidirectional Mamba design recovers accuracy approaching ViT-S but at 3--4$\times$ the energy cost.
TinyViT is Pareto-dominated (lower accuracy than ViT-S, higher energy than MambaVision-T).
On the smaller DTD dataset, architectural differences narrow---MambaVision-T LoRA reaches 0.772 vs ViT-S AdaLoRA at 0.778---suggesting that Vim-S's accuracy advantage is dataset-size dependent.
}

{%
\section{Gradient Checkpointing Analysis (Q2)}
\label{sec:ckpt-analysis}

This section examines how gradient checkpointing affects VRAM memory usage, energy consumption, and training time---the core dimensions of the memory--compute trade-off.

\subsection{Memory Reduction}
\label{sec:memory-reduction}

Table~\ref{tab:memory-reduction} summarizes \emph{peak} VRAM under checkpointing and the percentage reduction relative to the peak no-checkpointing baseline.
NO-CP is the per-configuration no-checkpointing baseline (peak VRAM without activation recomputation); $\Delta$ is the reduction relative to NO-CP. The baseline differs per (model, method) because freezing layers (LoRA/QLoRA/AdaLoRA/BitFit) avoids storing gradient-related activations.
Note that Table~\ref{tab:peft-cifar} reports \emph{average} VRAM for ViT-S/TinyViT, so the baseline values here differ.
Vim-S Full-FT without checkpointing is OOM at the 2\,GB budget, so its NO-CP and $\Delta$ cells are left as ``---''.

\begin{narrowtable}[htbp]
\centering
\caption{{Peak VRAM (MiB) under checkpointing on CIFAR-100 and  DTD.}
}
\label{tab:memory-reduction}
\footnotesize
\begin{tabular}{@{}llrrrrr@{}}
\toprule
\textbf{Model} & \textbf{Method} & \textbf{\shortstack{NO-CP\\(MiB)}} & \textbf{\shortstack{Static\\(MiB)}} & \textbf{$\Delta$\,(\%)} & \textbf{\shortstack{Adapt.\\(MiB)}} & \textbf{$\Delta$\,(\%)} \\
\midrule
\multicolumn{7}{c}{\textbf{CIFAR-100}} \\
\midrule
\multirow{5}{*}{ViT-S}
 & Full-FT  & 1{,}944 &  414    & $-$79 & 1{,}110 & $-$43 \\
 & LoRA     & 1{,}857 &  244    & $-$87 & 1{,}074 & $-$42 \\
 & AdaLoRA  & 1{,}800 &  258    & $-$86 & 1{,}089 & $-$39 \\
 & QLoRA    & 1{,}766 &  175    & $-$91 & 1{,}147 & $-$35 \\
 & BitFit   & 1{,}325 &  239    & $-$82 & 1{,}072 & $-$19 \\
\midrule
\multirow{5}{*}{TinyViT}
 & Full-FT  & 4{,}855 & 1{,}676 & $-$65 & OOM     & ---   \\
 & LoRA     & 2{,}170 & 1{,}192 & $-$45 & 1{,}241 & $-$43 \\
 & AdaLoRA  & 2{,}180 & 1{,}206 & $-$45 & 1{,}256 & $-$42 \\
 & QLoRA    & 2{,}122 & 1{,}129 & $-$47 & 1{,}176 & $-$45 \\
 & BitFit   & 4{,}160 & 1{,}197 & $-$71 & 3{,}057 & $-$27 \\
\midrule
\multirow{5}{*}{\shortstack[l]{Mamba-\\Vision-T}}
 & Full-FT  & 2{,}223 & 1{,}115 & $-$50 & 1{,}240 & $-$44 \\
 & LoRA     & 1{,}336 &  306    & $-$77 & 1{,}048 & $-$22 \\
 & AdaLoRA  & 1{,}352 &  328    & $-$76 & 1{,}062 & $-$21 \\
 & QLoRA    & 1{,}253 &  230    & $-$82 & 1{,}066 & $-$15 \\
 & BitFit   & 1{,}458 &  858    & $-$41 & 1{,}137 & $-$22 \\
\midrule
\multirow{5}{*}{Vim-S}
 & Full-FT  & OOM     &  578    & ---   & 1{,}438 & ---   \\
 & LoRA     & 6{,}618 &  404    & $-$94 & 1{,}474 & $-$78 \\
 & AdaLoRA  & 6{,}619 &  409    & $-$94 & 1{,}471 & $-$78 \\
 & QLoRA    & 6{,}538 &  323    & $-$95 & 1{,}394 & $-$79 \\
 & BitFit   & 5{,}230 &  377    & $-$93 & 1{,}217 & $-$77 \\
\midrule
\multicolumn{7}{c}{\textbf{DTD}} \\
\midrule
\multirow{5}{*}{ViT-S}
 & Full-FT  & 1{,}946 &  414    & $-$79 & 1{,}110 & $-$43 \\
 & LoRA     & 1{,}774 &  244    & $-$86 & 1{,}074 & $-$39 \\
 & AdaLoRA  & 1{,}781 &  252    & $-$86 & 1{,}079 & $-$39 \\
 & QLoRA    & 1{,}703 &  174    & $-$90 & 1{,}144 & $-$33 \\
 & BitFit   & 1{,}160 &  239    & $-$79 & 1{,}072 & $-$8  \\
\midrule
\multirow{5}{*}{TinyViT}
 & Full-FT  & 4{,}858 & 1{,}676 & $-$66 & OOM     & ---   \\
 & LoRA     & 2{,}097 & 1{,}191 & $-$43 & 1{,}241 & $-$41 \\
 & AdaLoRA  & 2{,}108 & 1{,}196 & $-$43 & 1{,}245 & $-$41 \\
 & QLoRA    & 2{,}030 & 1{,}125 & $-$45 & 1{,}175 & $-$42 \\
 & BitFit   & 4{,}109 & 1{,}197 & $-$71 & 3{,}041 & $-$26 \\
\midrule
\multirow{5}{*}{\shortstack[l]{Mamba-\\Vision-T}}
 & Full-FT  & 2{,}217 & 1{,}122 & $-$49 & 1{,}242 & $-$44 \\
 & LoRA     & 1{,}337 &  302    & $-$77 & 1{,}059 & $-$21 \\
 & AdaLoRA  & 1{,}343 &  306    & $-$77 & 1{,}065 & $-$21 \\
 & QLoRA    & 1{,}244 &  213    & $-$83 & 1{,}075 & $-$14 \\
 & BitFit   & 1{,}452 &  819    & $-$44 & 1{,}139 & $-$22 \\
\midrule
\multirow{5}{*}{Vim-S}
 & Full-FT  & OOM     &  579    & ---   & 1{,}439 & ---   \\
 & LoRA     & 6{,}618 &  404    & $-$94 & 1{,}477 & $-$78 \\
 & AdaLoRA  & 6{,}602 &  393    & $-$94 & 1{,}456 & $-$78 \\
 & QLoRA    & 6{,}537 &  323    & $-$95 & 1{,}394 & $-$79 \\
 & BitFit   & 5{,}230 &  376    & $-$93 & 1{,}212 & $-$77 \\
\bottomrule
\end{tabular}
\end{narrowtable}

Static checkpointing achieves the largest memory reductions: 79--91\% on ViT-S, 45--71\% on TinyViT, 41--82\% on MambaVision-T, and 93--95\% on Vim-S.

Vim-S achieves the greatest reduction in memory consumption because of its 24 Mamba blocks that accumulate 5--6.5\,GB of VRAM activations without checkpointing. A static checkpoint reduces this number to 323--578 \,MiB, which fits into a 2 \,GB budget.
Remarkably, Vim-S Full-FT causes OOM without a checkpoint, but with a checkpointing it uses only 578 \,MiB. It follows that the checkpoint is mandatory on this architecture in conditions of limited memory. On MambaVision-T, static checkpointing uniformly reduces memory consumption on all PEFT methods. 
For example: 77\% reduction in VRAM for LoRA (1{,}336$\to$306\,MiB), 82\% for QLoRA (1{,}253$\to$230\,MiB), 50\% for Full-FT (2{,}223$\to$1{,}115\,MiB), and 41\% for BitFit (1{,}458$\to$858\,MiB).  Full-FT and BitFit reflects the smallest reduction in memory consumption due to a larger proportion of the unchecked state (gradients of the full set of parameters and optimizer moments), rather than an architectural limitation.

Adaptive checkpoint shows more moderate reductions: 19--43\% on ViT-S, 27--45\% on TinyViT, 77--79\% on Vim-S. The main reason is that only the layers below the specified memory threshold are being checkpointed. On the ViT-S, the adaptive checkpoint is stably activated on Layer~7: Layers 1--6 store fully activated (${\sim}$124\,MiB per layer), and layers 7--12 are monitored (${\sim}$8\,MiB per layer). This boundary has been constant for 10 epochs, indicating the fact that the adaptive checkpoint algorithm converges rapidly and does not change.

\textbf{DTD.} The DTD block of Table~\ref{tab:memory-reduction} mirrors CIFAR-100, as expected: peak memory is fixed by architecture, batch size, and input resolution ($224{\times}224$, bs~32), not by training-set size, so the static and adaptive reduction ranges match their CIFAR-100 counterparts within a few MiB. The only differences---up to ${\sim}165$\,MiB on some NO-CP freezing-method rows---are run-to-run activation-measurement noise, orthogonal to the fine-tuning energy differences reported in Section~\ref{sec:timing-overhead}.

\begin{figure}[htbp]
    \centering
    \includegraphics[width=\textwidth]{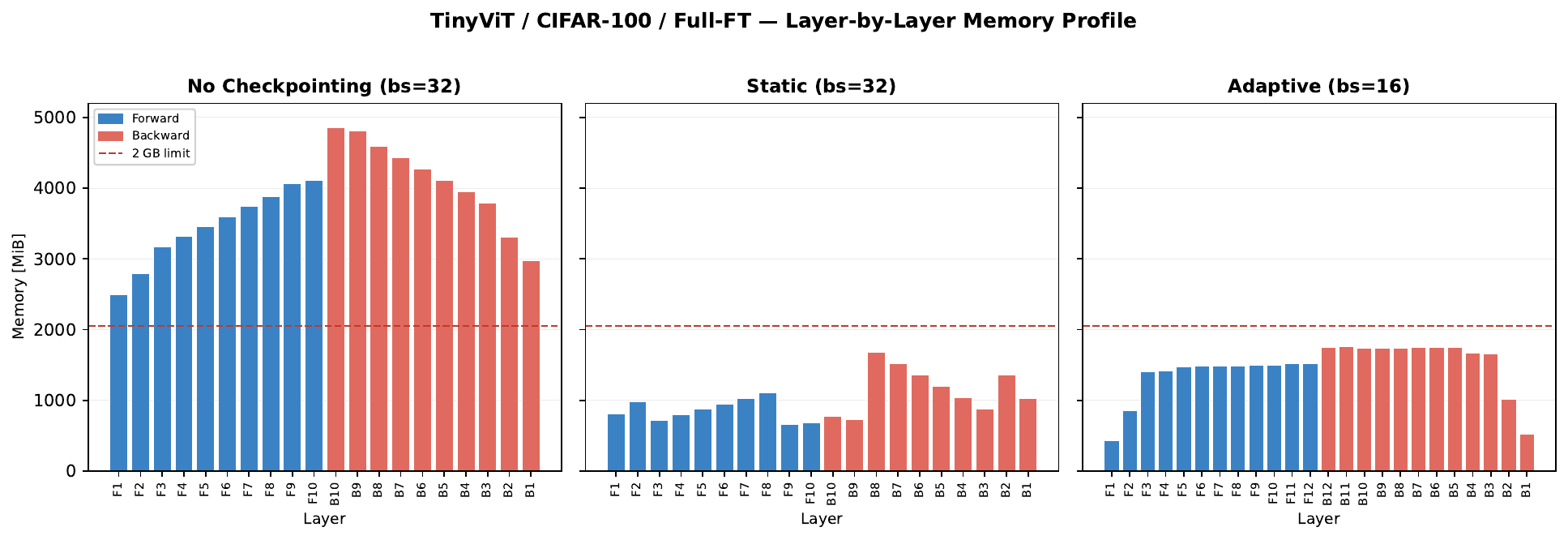}
    \caption{Layer-by-layer memory profile for TinyViT / CIFAR-100 / Full-FT at epoch~10. Left: no checkpointing at batch size~32 (peak 4855\,MiB). Center: static checkpointing at batch size~32 (peak 1676\,MiB). Right: adaptive checkpointing at batch size~16 (peak 1760\,MiB); batch size was halved because Full-FT adaptive triggers OOM at batch~32 on TinyViT's heterogeneous MBConv stem. The red dashed line marks the 2\,GB budget. 
    }
    \label{fig:waterfall-tinyvit}
\end{figure}

\subsection{Layer-by-Layer Memory Profiles}
\label{sec:memory-profiles}

The mechanism is visible in the per-layer memory profile: ViT-S grows linearly at ${\sim}124$\,MiB per Transformer layer, so checkpointing cuts its peak from $1{,}944$\,MiB to $414$ (static) or $1{,}110$\,MiB (adaptive, a $43\%$ reduction), whereas TinyViT's MBConv stem allocates $2{,}498$\,MiB in Layer~1 alone, producing the stepped profile in Fig.~\ref{fig:waterfall-tinyvit}. Because this breaks the uniform-growth assumption behind the adaptive heuristic, adaptive Full-FT on TinyViT requires bs\,=\,16 to fit the $2$\,GB budget, and static checkpointing remains the recommended default for heterogeneous CNN--Transformer hybrids.
\subsection{Fine-Tuning Energy and Time Increase under Checkpointing}
\label{sec:timing-overhead}


Table~\ref{tab:ckpt-overhead} reports absolute fine-tuning energy (Wh) for all four models across all five PEFT methods on both CIFAR-100 and DTD under the three checkpointing regimes, together with the \% increase relative to the per-configuration no-checkpointing baseline. 

\begin{narrowtable}[htbp]
\centering
\caption{Fine-tuning energy (Wh) under checkpointing on CIFAR-100 and DTD. NO-CP is the per-configuration no-checkpointing baseline; $\Delta$ is the \% increase relative to NO-CP. Vim-S Full-FT NO-CP is OOM; TinyViT Full-FT adaptive is OOM.
}
\label{tab:ckpt-overhead}
\footnotesize
\begin{tabular}{@{}llrrrrr@{}}
\toprule
\textbf{Model} & \textbf{Method} & \textbf{\shortstack{NO-CP\\(Wh)}} & \textbf{\shortstack{Static\\(Wh)}} & \textbf{$\Delta$\,(\%)} & \textbf{\shortstack{Adapt.\\(Wh)}} & \textbf{$\Delta$\,(\%)} \\
\midrule
\multicolumn{7}{c}{\textbf{CIFAR-100}} \\
\midrule
\multirow{5}{*}{ViT-S}
 & Full-FT  &  83.19 &  97.73 & $+$17 & 102.75 & $+$24 \\
 & LoRA     &  79.20 & 109.98 & $+$39 & 103.17 & $+$30 \\
 & AdaLoRA  &  92.25 & 126.59 & $+$37 & 118.29 & $+$28 \\
 & QLoRA    &  59.84 &  82.57 & $+$38 &  77.22 & $+$29 \\
 & BitFit   &  63.85 &  82.71 & $+$30 &  69.47 & $+$9  \\
\midrule
\multirow{5}{*}{TinyViT}
 & Full-FT  & 108.65 & 144.51 & $+$33 & OOM    & ---   \\
 & LoRA     &  87.13 & 115.84 & $+$33 & 124.25 & $+$43 \\
 & AdaLoRA  &  97.86 & 129.27 & $+$32 & 137.96 & $+$41 \\
 & QLoRA    &  74.00 & 101.42 & $+$37 & 109.81 & $+$48 \\
 & BitFit   &  84.43 & 115.68 & $+$37 & 144.15 & $+$71 \\
\midrule
\multirow{5}{*}{\shortstack[l]{Mamba-\\Vision-T}}
 & Full-FT  &  76.12 &  91.02 & $+$20 &  99.38 & $+$31 \\
 & LoRA     &  60.60 &  79.37 & $+$31 &  72.94 & $+$20 \\
 & AdaLoRA  &  66.74 &  85.51 & $+$28 &  79.03 & $+$18 \\
 & QLoRA    &  54.56 &  66.71 & $+$22 &  61.85 & $+$13 \\
 & BitFit   &  58.01 &  72.31 & $+$25 &  71.83 & $+$24 \\
\midrule
\multirow{5}{*}{Vim-S}
 & Full-FT  & OOM    & 308.51 & ---   & 306.57 & ---   \\
 & LoRA     & 239.24 & 331.56 & $+$39 & 316.97 & $+$32 \\
 & AdaLoRA  & 252.01 & 327.38 & $+$30 & 329.08 & $+$31 \\
 & QLoRA    & 253.95 & 327.33 & $+$29 & 338.51 & $+$33 \\
 & BitFit   & 205.27 & 277.33 & $+$35 & 271.53 & $+$32 \\
\midrule
\multicolumn{7}{c}{\textbf{DTD}} \\
\midrule
\multirow{5}{*}{ViT-S}
 & Full-FT  &  71.63 &  89.25 & $+$25 &  93.61 & $+$31 \\
 & LoRA     &  68.45 &  98.74 & $+$44 &  92.07 & $+$35 \\
 & AdaLoRA  &  80.29 & 113.99 & $+$42 & 108.79 & $+$35 \\
 & QLoRA    &  53.91 &  72.99 & $+$35 &  68.32 & $+$27 \\
 & BitFit   &  56.20 &  74.21 & $+$32 &  61.37 & $+$9  \\
\midrule
\multirow{5}{*}{TinyViT}
 & Full-FT  &  97.48 & 134.75 & $+$38 & OOM    & ---   \\
 & LoRA     &  74.88 & 104.98 & $+$40 & 110.91 & $+$48 \\
 & AdaLoRA  &  87.81 & 119.27 & $+$36 & 127.89 & $+$46 \\
 & QLoRA    &  65.52 &  90.93 & $+$39 & 101.26 & $+$55 \\
 & BitFit   &  74.23 & 105.60 & $+$42 & 128.77 & $+$73 \\
\midrule
\multirow{5}{*}{\shortstack[l]{Mamba-\\Vision-T}}
 & Full-FT  &  67.88 &  81.92 & $+$21 &  91.25 & $+$34 \\
 & LoRA     &  54.49 &  70.61 & $+$30 &  64.78 & $+$19 \\
 & AdaLoRA  &  61.05 &  78.92 & $+$29 &  73.64 & $+$21 \\
 & QLoRA    &  50.50 &  61.65 & $+$22 &  56.69 & $+$12 \\
 & BitFit   &  52.40 &  65.32 & $+$25 &  83.62 & $+$60 \\
\midrule
\multirow{5}{*}{Vim-S}
 & Full-FT  & OOM    & 277.39 & ---   & 340.17 & ---   \\
 & LoRA     & 218.74 & 288.58 & $+$32 & 293.81 & $+$34 \\
 & AdaLoRA  & 221.57 & 300.59 & $+$36 & 309.16 & $+$40 \\
 & QLoRA    & 223.09 & 300.77 & $+$35 & 312.58 & $+$40 \\
 & BitFit   & 185.64 & 248.74 & $+$34 & 253.76 & $+$37 \\
\bottomrule
\end{tabular}
\end{narrowtable}


\textbf{CIFAR-100.}
Static checkpointing increases fine-tuning energy consumption by 17--39\% on ViT-S and 32--37\% on TinyViT relative to the no-checkpointing baseline.
This increase is the direct cost of activation recomputation: every layer's forward pass is executed twice (once in the forward pass, once during the backward pass), approximately doubling the forward-pass computation time. The resulting fine-tuning time increase is not exactly $2\times$ because the forward pass is typically faster than the backward pass, and additional operations (gradient accumulation, optimizer step) are unaffected by checkpointing.

Adaptive checkpointing imposes a smaller fine-tuning energy increase on ViT-S (9--30\% depending on the PEFT method) because it only recomputes the checkpointed layers (Layers 7--12, i.e., 50\% of layers). BitFit with adaptive checkpointing on ViT-S consumes only 9\% more energy (69.47\,Wh vs 63.85\,Wh), making it the most practical memory-saving configuration for this architecture.
On TinyViT, however, adaptive checkpointing consumes \emph{more} fine-tuning energy than static for several methods. BitFit adaptive consumes 144.15\,Wh versus 115.68\,Wh for static---a 25\% additional increase. 
The per-layer timing explains why: on TinyViT the adaptive heuristic produces 12 layers instead of the expected 10 and inflates forward and backward time by $45\%$ and $100\%$ respectively, because a splitting rule tuned for uniform Transformer blocks partitions the heterogeneous MBConv+Attention stack poorly.

\textbf{DTD.}
\textbf{DTD.} DTD shows the same qualitative pattern: static checkpointing is broadly usable across architectures, while adaptive is best on Transformer backbones and undesirable on heterogeneous ones. On ViT-S static costs $25$--$44\%$ extra energy (vs.\ $17$--$39\%$ on CIFAR-100) and adaptive $27$--$35\%$ (vs.\ $9$--$30\%$); the smaller dataset shifts the balance toward backward-pass cost and erodes adaptive's advantage. On TinyViT static stays near the CIFAR-100 range ($36$--$42\%$) but adaptive is far worse ($46$--$73\%$ vs.\ at most $25\%$ on CIFAR-100), confirming that adaptive's suboptimal MBConv partitioning is amplified, not masked, on a new dataset.


\textbf{Key Findings on Checkpointing}:
%
The right checkpointing strategy is dictated by the backbone, not by the PEFT method:
1)~on pure-Mamba backbones (Vim-S) checkpointing is non-negotiable---full fine-tuning is infeasible without it---so static is the sensible default and adaptive adds little value;
2)~on uniform Transformer backbones (ViT-S) adaptive is preferable, retaining most of static's memory saving while roughly halving the fine-tuning energy overhead;
3)~on heterogeneous hybrids (TinyViT, MBConv$+$Attention) adaptive should be avoided---a single memory threshold misallocates checkpoints across stages whose footprints differ by an order of magnitude and can exceed even static's energy cost, so fall back to static and treat stage-aware adaptation as future work.

\begin{table}[htbp]
\centering
\caption{Contrastive and KD-SSL baseline results (no training). SigLIP and OpenCLIP use text-guided zero-shot classification; DINOv2 uses $k$-NN and linear probe on frozen features (time/energy include feature extraction). Params: total model parameters (vision$+$text for contrastive; vision-only for DINOv2). The four rightmost columns are the $1/8$-weight deployment-only NetScore subset (Section~\ref{sec:ns-metric}, Table~\ref{tab:ns-variants}). \textbf{Bold}: best per dataset section.
}
\label{tab:contrastive_results}
\footnotesize
\resizebox{\textwidth}{!}{%
\begin{tabular}{ll r >{}r r r r r >{}r>{}r>{}r>{}r}
\toprule
\textbf{Model} & \textbf{Method} & \textbf{\shortstack{Params\\(M)}} & {\textbf{\shortstack{FLOPs\\(G)}}} & \textbf{Acc} & \textbf{\shortstack{Inf. Time\\(min)}} & \textbf{\shortstack{Avg Pwr\\(W)}} & \textbf{\shortstack{Energy\\(Wh)}} & {\textbf{NS}} & {\textbf{\shortstack{NS$_{E}$\\$1/8$}}} & {\textbf{\shortstack{NS$_{M}$\\$1/8$}}} & {\textbf{\shortstack{NS$^{\#}$\\$1/8$}}} \\
\midrule
\multicolumn{12}{c}{\textbf{CIFAR-100}} \\
\midrule
SigLIP-Large      & Zero-shot & 652 & 190.7 & 0.808 & 3.65  & 164.2 & 9.92  & 25.35 & 64.91 & 68.22 & 56.82 \\
OpenCLIP ViT-L/14 & Zero-shot & 428 & 81.1  & 0.873 & \textbf{1.30}  & 167.8 & \textbf{3.19}  & 32.24 & \textbf{67.35} & 70.09 & \textbf{59.80} \\
DINOv2-Large      & $k$-NN    & 304 & 81.1  & 0.912 & 14.34 & 149.5 & 35.45 & 34.48 & 65.63 & 71.11 & 58.35 \\
DINOv2-Large      & Linear    & 304 & 81.1  & \textbf{0.917} & 14.43 & 147.6 & 35.50 & \textbf{34.58} & 65.73 & \textbf{71.21} & 58.44 \\
\midrule
\multicolumn{12}{c}{\textbf{DTD Textures}} \\
\midrule
SigLIP-Large      & Zero-shot & 652 & 190.7 & 0.710 & 0.88  & 130.2 & 1.59  & 23.10 & 64.46 & 65.97 & 56.38 \\
OpenCLIP ViT-L/14 & Zero-shot & 428 & 81.1  & 0.678 & \textbf{0.43}  & 122.1 & \textbf{0.44}  & 27.84 & 64.50 & 65.71 & 56.96 \\
DINOv2-Large      & $k$-NN    & 304 & 81.1  & \textbf{0.740} & 1.27  & 126.5 & 2.08  & \textbf{30.85} & \textbf{64.81} & \textbf{67.49} & \textbf{57.52} \\
DINOv2-Large      & Linear    & 304 & 81.1  & 0.728 & 1.28  & 97.9  & 2.08  & 30.57 & 64.79 & 67.20 & 57.51 \\
\bottomrule
\end{tabular}%
}
\end{table}

\begin{table}[htbp]
\centering
\caption{VLM zero-shot and QLoRA fine-tuning results. ZS rows: single-prompt classification, no training. QLoRA rows: light-touch all-layer LoRA ($r{=}16$, $\alpha{=}32$), 1 epoch. \textbf{FT Time / FT E}: one-time training cost. \textbf{Inf.\ Time / Avg Pwr / Inf.\ E}: single test-set pass (comparable across ZS and QLoRA). Params: total model parameters. Four rightmost columns are the $1/8$-weight deployment-only NetScore subset (Section~\ref{sec:ns-metric}, Table~\ref{tab:ns-variants}). \textbf{Bold}: best per dataset section. \textsuperscript{$\ddagger$}NS uses analytical per-image FLOPs (±20\%). \textsuperscript{$\dagger$}SmolVLM ZS energy monitoring failed (\texttt{pynvml} unavailable). The MobileVLM-DTD and PaliGemma QLoRA rows use the unified $800$-step schedule (lr $5{\times}10^{-5}$, patience $5$); the default 300-step recipe early-stopped at the ZS plateau.
}
\label{tab:vlm_results}
\footnotesize
\resizebox{\textwidth}{!}{%
\begin{tabular}{ll r r r r r r r >{}r>{}r>{}r>{}r}
\toprule
\textbf{Model} & \textbf{Method} & \textbf{\shortstack{Params\\(M)}} & \textbf{Acc} & \textbf{\shortstack{FT Time\\(min)}} & \textbf{\shortstack{FT E\\(Wh)}} & \textbf{\shortstack{Inf. Time\\(min)}} & \textbf{\shortstack{Avg Pwr\\(W)}} & \textbf{\shortstack{Inf. E\\(Wh)}} & {\textbf{NS}} & {\textbf{\shortstack{NS$_{E}$\\$1/8$}}} & {\textbf{\shortstack{NS$_{M}$\\$1/8$}}} & {\textbf{\shortstack{NS$^{\#}$\\$1/8$}}} \\
\midrule
\multicolumn{13}{c}{\textbf{CIFAR-100}} \\
\midrule
SmolVLM-256M      & Zero-shot & 256     & 0.135           & ---            & ---            & 98.48           & ---\textsuperscript{$\dagger$} & ---\textsuperscript{$\dagger$} & 0.60\textsuperscript{$\ddagger$} & ---\textsuperscript{$\dagger$} & 37.74 & ---\textsuperscript{$\dagger$} \\
SmolVLM-256M      & QLoRA     & 256     & 0.397           & 38.41          & 36.93          & 106.33          & 168.4 & 298.43          & \textbf{19.34}\textsuperscript{$\ddagger$} & 48.87          & 56.43 & 41.34 \\
MobileVLM V2-1.7B & Zero-shot & 1{,}700 & 0.274           & ---            & ---            & 45.69           & 73.7  & 55.73           & $-5.76$\textsuperscript{$\ddagger$} & 44.25          & 48.70 & 35.44 \\
MobileVLM V2-1.7B & QLoRA     & 1{,}700 & \textbf{0.517}  & \textbf{11.50} & \textbf{3.34}  & 43.38           & 124.0 & 89.68           & 5.27\textsuperscript{$\ddagger$} & \textbf{54.77} & \textbf{60.43} & \textbf{46.65} \\
PaliGemma-3B      & Zero-shot & 3{,}030 & 0.229           & ---            & ---            & \textbf{32.54}  & 83.8  & \textbf{45.04}  & $-13.67$\textsuperscript{$\ddagger$} & 41.36          & 44.89 & 31.85 \\
PaliGemma-3B      & QLoRA     & 3{,}030 & 0.454           & 27.88          & 6.02           & 88.46           & 126.0 & 185.77          & $-1.77$\textsuperscript{$\ddagger$} & 51.72          & 57.61 & 43.05 \\
\midrule
\multicolumn{13}{c}{\textbf{DTD Textures}} \\
\midrule
SmolVLM-256M      & Zero-shot & 256     & 0.078           & ---            & ---            & 19.55           & ---\textsuperscript{$\dagger$} & ---\textsuperscript{$\dagger$} & $-7.64$\textsuperscript{$\ddagger$} & ---\textsuperscript{$\dagger$} & 28.21 & ---\textsuperscript{$\dagger$} \\
SmolVLM-256M      & QLoRA     & 256     & 0.390           & 53.70          & 50.12          & 20.59           & 155.9 & 53.52           & \textbf{20.32}\textsuperscript{$\ddagger$} & 50.43          & 56.11 & 42.91 \\
MobileVLM V2-1.7B & Zero-shot & 1{,}700 & 0.040           & ---            & ---            & 14.57           & 66.5  & 16.09           & $-38.00$\textsuperscript{$\ddagger$} & 12.17          & 15.27 & 3.36  \\
MobileVLM V2-1.7B & QLoRA     & 1{,}700 & \textbf{0.528}  & 25.93          & 5.65           & 12.09           & 58.8  & 11.84           & 6.82\textsuperscript{$\ddagger$} & \textbf{57.33} & \textbf{60.79} & \textbf{49.22} \\
PaliGemma-3B      & Zero-shot & 3{,}030 & 0.215           & ---            & ---            & \textbf{5.53}   & 83.6  & \textbf{7.61}   & $-13.75$\textsuperscript{$\ddagger$} & 42.19          & 43.79 & 32.69 \\
PaliGemma-3B      & QLoRA     & 3{,}030 & 0.488           & \textbf{25.89} & \textbf{5.30}  & 12.44           & 120.1 & 24.90           & 0.51\textsuperscript{$\ddagger$} & 55.16          & 58.88 & 46.50 \\
\bottomrule
\end{tabular}%
}
\end{table}

{%
\section{Learning Paradigm Comparison (Q3)}
\label{sec:paradigm-comparison}

This section compares transfer learning (PEFT fine-tuning) against training-free foundation-model baselines across three paradigms: contrastive learning, knowledge-distillation self-supervised learning (KD-SSL), and autoregressive vision-language models.

\subsection{Contrastive and KD-SSL Baselines}
\label{sec:results-contrastive}

Table~\ref{tab:contrastive_results} presents the baseline results for two contrastive models and one KD-SSL model. SigLIP and OpenCLIP perform text-guided zero-shot classification, while DINOv2---a knowledge-distillation self-supervised model---is evaluated via $k$-NN retrieval and linear probe on frozen features. Following the comparison protocol established below, all numbers in this section compare \emph{inference} cost against \emph{inference} cost: the Inf.~Time and Inf.~E columns of Tables~\ref{tab:peft-cifar}--\ref{tab:peft-dtd} are the fine-tuned counterpart to the Energy column of Table~\ref{tab:contrastive_results}.

DINOv2-Large (KD-SSL) achieves the highest accuracy on both datasets ($0.917$ on CIFAR-100, $0.740$ on DTD); on CIFAR-100 its linear probe even surpasses the best fine-tuned model (ViT-S Full-FT, $0.897$). It is, however, far more expensive at inference, since feature extraction runs the full ViT-L/14 forward pass per image ($35.50$\,Wh on CIFAR-100, $2.08$\,Wh on DTD). Inference-for-inference, DINOv2 Linear costs $35.50$\,Wh against $0.759$\,Wh for ViT-S QLoRA ($0.876$) or $0.782$\,Wh for MambaVision-T QLoRA ($0.835$)---about $45\times$ more energy for a $0.02$--$0.08$ accuracy gain. Even the one-time training of ViT-S QLoRA ($59.84$\,Wh over 10 epochs) is only ${\sim}1.5\times$ a single DINOv2 pass, after which every fine-tuned inference draws ${\sim}45\times$ less energy. DINOv2 is thus the right choice only when accuracy outweighs the inference budget and training is not an option; otherwise the fine-tuned route wins as soon as inference repeats.
%

{By the NetScore columns of Table~\ref{tab:contrastive_results}, the complexity-aware NS goes to the highest-accuracy KD-SSL variant (DINOv2 Linear $34.58$ on CIFAR-100, $k$-NN $30.85$ on DTD), while the inference-cost-aware NS$_{E}$ favors OpenCLIP (CIFAR-100 $67.35$; on DTD the leader depends on penalty strength---DINOv2 $k$-NN at $1/8$, OpenCLIP at $1/4$, rewarding its ${\sim}3\times$ shorter inference).} The inference-for-inference picture is more nuanced than a single metric suggests: on CIFAR-100 OpenCLIP reaches $0.873$ at $3.19$\,Wh---competitive with DINOv2 on energy but \emph{above} every fine-tuned ViT-S/MambaVision-T PEFT configuration ($0.759$--$1.437$\,Wh), while trailing ViT-S QLoRA ($0.876$) and ViT-S Full-FT ($0.897$) on accuracy.
OpenCLIP is therefore attractive when no training budget exists or broad label coverage is needed, but it is not Pareto-dominant once fine-tuned models are compared on inference cost alone.

On DTD, the ordering flips in favor of fine-tuning: the best fine-tuned result (ViT-S AdaLoRA, 0.778 at 0.222\,Wh inference) exceeds the best baseline result (DINOv2 $k$-NN, 0.740 at $2.08$\,Wh inference) on both accuracy and inference energy.
OpenCLIP (0.678 at $0.44$\,Wh) trails PEFT methods even on the cheaper inference axis at this fine-grained texture domain. The training cost of PEFT ($68$--$80$\,Wh for ViT-S AdaLoRA/LoRA) is a one-time investment amortized across all subsequent inference requests.

{%
\subsection{VLM Baselines}
\label{sec:results-vlm}

Table~\ref{tab:vlm_results} presents the results for three vision-language models evaluated via zero-shot prompting. Each model processes the test images by generating text responses to classification prompts, with no training or feature adaptation.
The VLM baselines achieve substantially lower accuracy than both contrastive models and fine-tuned transfer learning approaches. 
MobileVLM V2-1.7B is the best-performing VLM on CIFAR-100 at 0.274, while PaliGemma-3B leads on DTD at 0.215. All three VLMs achieve very low accuracy on DTD (MobileVLM 0.040, SmolVLM 0.078), only marginally above the $1/47 \approx 0.021$ random baseline.



{A causal analysis of this gap, together with mitigations such as instruction-style fine-tuning of the decoder, is deferred to Section~\ref{sec:limitations} (Discussion).}

Comparing inference against inference makes the gap starker. On CIFAR-100 MobileVLM and PaliGemma draw $55.73$ and $45.04$\,Wh per pass, whereas the Pareto-competitive ViT-S QLoRA costs only $0.759$\,Wh ($60$--$73\times$ less); even the most expensive fine-tuned inference (Vim-S QLoRA, $3.408$\,Wh) is an order of magnitude cheaper. A single VLM evaluation pass exceeds a full 10-epoch training of several PEFT configurations (e.g., ViT-S QLoRA trains in $59.84$\,Wh, Table~\ref{tab:peft-cifar}). VLMs are thus uncompetitive on both axes---lower accuracy \emph{and} higher inference energy---and cannot substitute for contrastive, KD-SSL, or PEFT approaches on standard classification.

Although SmolVLM's power, energy, and NS$_{E}$ are missing (Table~\ref{tab:vlm_results}, $\dagger$), its ranking can be bounded. On CIFAR-100 it has the lowest accuracy ($0.135$) and the longest inference time ($98.48$\,min), so at a power comparable to the other VLMs its inference energy (${\sim}120$--$130$\,Wh) and NS$_{E}$ would be the worst on every axis. On DTD it is not the accuracy floor ($0.078$ vs.\ MobileVLM $0.040$), but its inference time ($19.55$\,min) is again the longest, so its energy and NS$_{E}$ remain among the worst of the zero-shot VLMs.

{
\textbf{Approach 2 (class-definition prompts): per-pair numerical breakdown.}
The mitigation experiment discussed in Section~\ref{sec:limitations} (``Three mitigation approaches'') was evaluated as follows. On DTD, prepending $\sim$500-token class definitions collapsed PaliGemma-3B from $21.54\%$ to $1.28\%$ ($-20.3$\,pp) and moved MobileVLM~V2-1.7B from $3.99\%$ to only $4.68\%$ ($+0.7$\,pp, within noise). On CIFAR-100, hierarchical two-step prompting ($20$ superclasses $\to$ $5$ leaves, two generations per image) dropped MobileVLM from $27.40\%$ to $11.94\%$ ($-15.5$\,pp; the superclass step reaches only $23.64\%$, so errors compound), while PaliGemma failed to complete the hierarchical protocol and SmolVLM-256M could not load either variant. Inference cost moves in the wrong direction at the same time: the two-generation hierarchical pass raises MobileVLM CIFAR-100 evaluation to $81.1$\,min / $192.1$\,Wh (vs.\ $45.69$\,min / $55.73$\,Wh zero-shot, Table~\ref{tab:vlm_results}), and the $500$-token DTD definition prompts raise per-pass energy to $32.5$\,Wh (MobileVLM) and $42.9$\,Wh (PaliGemma).

\textbf{Approach 3 (lightweight QLoRA fine-tuning): per-pair gains, recipe, and inference cost.}
The QLoRA rows of Table~\ref{tab:vlm_results} report a single epoch of all-layer LoRA ($r{=}16$, $\alpha{=}32$) with answer-only loss masking and validation-based early stopping --- the smallest realisation of the data-side prescription discussed in Section~\ref{sec:limitations}, Approach~3. Per-pair accuracy gains relative to the zero-shot row of the same model are: PaliGemma-3B $+23$\,pp on CIFAR-100 and $+27$\,pp on DTD, MobileVLM~V2-1.7B $+24$\,pp on CIFAR-100 and $+49$\,pp on DTD, SmolVLM-256M $+26$\,pp on CIFAR-100 and $+31$\,pp on DTD. Two recipe observations stand out. (i) Schedule length matters more than recipe design when the zero-shot starting point is weak: under the default schedule ($300$ steps, lr $10^{-5}$, patience $3$) MobileVLM-DTD did not move ($0.040\to 0.039$) because validation accuracy plateaued at the near-random zero-shot level and early stopping fired before learning began, whereas the firmer schedule ($800$ steps, lr $5{\times}10^{-5}$, patience $5$) produced a monotonic validation curve ($5\%\to 54.5\%$, still climbing at the last step) and lifted final accuracy to $0.528$, the best VLM result on DTD. (ii) The fine-tuned MobileVLM also decodes \emph{faster and cheaper} than its zero-shot counterpart on DTD ($12.09$ vs.\ $14.57$\,min, $11.84$ vs.\ $16.09$\,Wh per test-set pass), because the adapter teaches the decoder to emit a short class name followed by end-of-sequence instead of free-form caption text. The inference-energy axis, however, does \emph{not} close: the fine-tuned VLMs consume $12$--$298$\,Wh per test-set pass, two to three orders of magnitude above the $0.76$\,Wh of ViT-S~+~QLoRA at $0.876$ CIFAR-100 accuracy (Table~\ref{tab:peft-cifar}) and above the $3.19$\,Wh of OpenCLIP at $0.873$ accuracy (Table~\ref{tab:contrastive_results}).
}
}



\textbf{Key Findings on Learning Paradigms}:
The right paradigm depends on how closely the target domain matches the upstream pretraining distribution and on whether a training budget is available: 
1) Knowledge-distilled self-supervised features (DINOv2) can match or exceed full fine-tuning on generic natural-image benchmarks without any gradient updates, making them the preferred default when the target domain is well-represented by the pretraining corpus and no training compute is available. 
2) Contrastive text-guided models (OpenCLIP, SigLIP) occupy the opposite end of the accuracy--energy trade-off: they deliver strong accuracy at a minimal inference cost because aligned image--text embeddings transfer well to broad categories, and they are a reasonable baseline whenever a small inference cost is acceptable and exact label coverage is not required. 
3) On fine-grained or out-of-distribution domains such as texture classification, all three foundation-model baseline paradigms fall short of PEFT fine-tuning; when the target task deviates materially from the pretraining distribution, paying the one-time PEFT cost remains the right choice.
Current small-scale autoregressive VLMs are not competitive for standard classification: their per-image text-generation overhead drives inference energy above the contrastive and KD-SSL baselines while delivering lower accuracy, so they should not be substituted for contrastive or fine-tuned approaches on well-defined label sets.
}

\section{Summary of Questions}
\label{sec:results-summary}

Table~\ref{tab:summary} consolidates the best configuration per question. Detailed per-question findings appear in the intermediate summaries of the previous sections. 

\begin{narrowtable}[htbp]
\centering
\caption{Best configurations per question on CIFAR-100.}
\label{tab:summary}
\footnotesize
\begin{tabular}{@{}p{1.0cm}p{2.8cm}p{3.5cm}@{}}
\toprule
\textbf{Question} & \textbf{Best config.} & \textbf{Key result} \\
\midrule
Q1 (PEFT)  & ViT-S + BitFit & NS\,=\,81.69 (best across all configs); $-$23\% energy vs Full-FT \\
Q1 (Arch.) & ViT-S (accuracy); MV-T (energy) & Mamba models not uniformly efficient \\
Q2 & ViT-S + adaptive & $-$43\% VRAM, +9--30\% energy \\
Q3 & DINOv2 linear probe & 0.917 acc at 35.5\,Wh \\
\bottomrule
\end{tabular}
\end{narrowtable}

Across all 80+ configurations, ViT-S with QLoRA or BitFit and no checkpointing offers the best overall accuracy--energy trade-off for unconstrained memory budgets. 
Under this constraint, adding static checkpointing retains this ranking at a 17--44\% additional energy cost (CIFAR-100 17--39\%, DTD 25--44\%).
%
When training is not required, DINOv2 (KD-SSL) dominates on CIFAR-100 accuracy (0.917) but at roughly $11\times$ the inference cost of the contrastive alternative ($14.43$\,min / $35.50$\,Wh vs.\ OpenCLIP $1.30$\,min / $3.19$\,Wh on CIFAR-100); OpenCLIP ViT-L/14 is therefore the cheaper-inference zero-shot choice when its $0.044$ accuracy gap to DINOv2 is acceptable, while DINOv2 is reserved for settings where that margin outweighs the higher inference time and energy. }
PEFT fine-tuning remains necessary for domain-specific datasets such as DTD where the foundation-model baselines fall short.

\chapter{Discussion}
\label{sec:discussion}
\label{sec:limitations}




{
\textbf{Why autoregressive VLMs underperform on standard classification.}
The low accuracy of MobileVLM, PaliGemma, and SmolVLM (Table~\ref{tab:vlm_results}) mirrors the systematic diagnosis offered by two recent studies of visually-grounded language models (VLMs) on closed-set image classification, Zhang~et~al.~\cite{zhang2024vlmclassification} and Epstein~et~al.~\cite{epstein2025rethinking}, which converge on two main findings.

First, the bottleneck is in \emph{decoding} rather than visual representation, prompt variation, and inference strategy: Zhang~et~al.~\cite{zhang2024vlmclassification} show that a linear probe on a VLM's frozen vision encoder recovers near-CLIP-level classification accuracy, while the same model asked to \emph{generate} the label underperforms by a wide margin, because open-vocabulary text generation allows synonyms, hypernyms, hedges, and off-vocabulary strings that all count as wrong even when the image is correctly recognised. This decoding mismatch is what separates our autoregressive baselines from the contrastive zero-shot models (OpenCLIP, DINOv2, SigLIP), which score the closed label set directly via image--text similarity~\cite{radford2021learning} and therefore never expose the failure mode above.

Second, the root cause is a \emph{data} problem rather than an architectural one: both~\cite{zhang2024vlmclassification} and~\cite{epstein2025rethinking} attribute the gap to the under-representation of classification-style supervision in the public instruction-tuning corpora used to train small VLMs, which leaves the decoder defaulting to caption-style outputs and yields weak vision-grounded class priors --- especially in fine-grained domains such as DTD. Both studies converge on the same prescription, namely targeted exposure to closed-vocabulary classification data via instruction-style fine-tuning or constrained decoding, which directly motivates the three mitigation experiments reported next.

Third, the decoding penalty is amplified by the sub-word tokenisation used in all three VLM stacks (BPE-style segmentation~\cite{sennrich2016bpe}, in practice via the SentencePiece tokenisers~\cite{kudo2018sentencepiece} shipped with the LLaMA-family and Gemma decoders behind MobileVLM, SmolVLM, and PaliGemma): many class names tokenise into multiple sub-word tokens (e.g.\ ``aquarium\_fish'', ``interlaced'', ``lacelike''), so reaching the correct label requires a low-probability joint sequence of token decisions, and the error probability compounds with sequence length on top of the two failure modes above.
}

{
\textbf{Three mitigation approaches tried, but only fine-tuning is effective.}
Building on the previous discussion, we evaluated three routes for minimizing the autoregressive-VLM accuracy gap, ordered from cheapest to most expensive in training cost: (1)~\emph{trie-constrained beam-search decoding}, (2)~\emph{class-definition prompt engineering}, and (3)~\emph{lightweight QLoRA fine-tuning} of the decoder. 

\textit{Approach 1: trie-constrained beam-search decoding.}
Trie-constrained beam search restricts the decoder's output distribution at every step to tokens that continue a valid label-name prefix, collapsing the open-vocabulary search space to the closed label set at zero training cost; therefore, it directly targets the decoding-vs-encoding mismatch identified by the diagnosis~\cite{zhang2024vlmclassification}. Despite eliminating off-vocabulary outputs by construction, trie decoding improves accuracy only marginally, confirming the data-side diagnosis: the gap is dominated by representation and class-prior quality rather than by decoding noise alone, so a pure decoding fix cannot recover the missing accuracy on its own.
\emph{Future direction:} the small accuracy lift makes trie decoding a low-cost add-on rather than a load-bearing fix; this paper will combine it with Approach~3 (QLoRA-FT) to test whether the two interventions are complementary, and will report the same trie-vs-baseline numbers on all three VLMs and both datasets.

\textit{Approach 2: class-definition prompt engineering.}
Approach~2 prepends a textual class description --- a long ($\sim$500-token) DTD definition or a two-step superclass-then-class CIFAR decomposition --- to strengthen the decoder's vision-grounded class prior, the second failure mode identified by~\cite{epstein2025rethinking}. The outcome lands in the worst-case quadrant of the accuracy--energy plane on every measured $\langle$model, dataset$\rangle$ pair (per-pair numbers reported in Section~\ref{sec:results-vlm}): accuracy drops while inference cost simultaneously rises, which is consistent with --- and \emph{even worse than} --- the prediction of~\cite{zhang2024vlmclassification,epstein2025rethinking} that prompt-only interventions cannot recover what is missing in the encoder representation and the decoder's class-prior data.
\emph{Future direction:} naive prompt elaboration is not the right axis, but shorter LLM-generated visual descriptors in the CuPL/Menon--Vondrick style~\cite{pratt2023cupl,menon2023visual} (typically $\le 50$ tokens per class) are known to add $1$--$6$\,pp on fine-grained benchmarks at negligible inference cost; reproducing this protocol on our exact VLM backbones is avaliable, with a strict per-prompt token budget to avoid the inference-cost blow-up seen above.

\textit{Approach 3: lightweight QLoRA fine-tuning.}
Approach~3 is the only intervention that produces a material accuracy gain: a single epoch of light-touch all-layer QLoRA closes most of the zero-shot accuracy gap on every measured pair (per-pair gains, recipe, and inference cost reported in Section~\ref{sec:results-vlm}). This is the smallest realisation of the data-side prescription on which both~\cite{zhang2024vlmclassification} and~\cite{epstein2025rethinking} converge, and the size of the gain empirically corroborates their diagnosis: most of the zero-shot gap is decoding mismatch rather than representation gap, since a small adapter that nudges the decoder toward the closed label set already recovers a large fraction of the missing accuracy. The inference-energy axis, however, does \emph{not} close: even after QLoRA the autoregressive VLM family stays two to three orders of magnitude above the cheapest fine-tuned Transformer and the cheapest contrastive zero-shot baseline on inference cost, so it becomes competitive with --- but remains Pareto-dominated by --- both transfer learning and contrastive ZS once inference cost is accounted for. The opening Discussion bullet (small autoregressive VLMs are not the right tool for standard closed-set classification under a deployment-energy budget) therefore survives the QLoRA experiment intact.

\emph{Future direction:} two complementary future works close the residual gaps from this experiment. \emph{(a) Per-pair schedule calibration:} longer-horizon training on configurations whose validation curve has not saturated (MobileVLM-DTD and SmolVLM-DTD were still climbing at the last step) and a per-pair search over $\{$steps, lr, patience$\}$ to push accuracy further before any new recipe is considered. \emph{(b) Inference-energy reduction:} vision-token compression at the projector (Q-Former, pruning, or merging) to cut the prefix length, speculative decoding with a small draft model to reduce per-token latency, and KV-cache quantisation/reuse across same-prompt batches to amortise the prefill phase --- together targeting the $90$--$400\times$ inference-energy gap that training-side mitigations alone cannot remove.

\chapter{Conclusion}
\label{ch:conclusion}

This study benchmarked five PEFT methods across Transformer (ViT-S, TinyViT), pure Mamba (Vim-S), and Mamba-hybrid (MambaVision-T) architectures under an on-device VRAM budget (here 2\,GB) on CIFAR-100 and DTD, complemented by zero-shot VLM baselines and a novel adaptive checkpointing algorithm. 
QLoRA and BitFit consistently deliver the best accuracy--energy trade-off, achieving accuracy within 0.01--0.02 of Full Fine-Tuning at 20--30\% lower energy cost on ViT-S. 
The architectural comparison reveals that ViT-S leads in accuracy (0.897) while MambaVision-T is the most energy-efficient (8--28\% less energy), with TinyViT dominated on both axes. 
Vim-S recovers accuracy close to ViT-S (0.872) but at 3--4$\times$ the energy cost, demonstrating that Mamba architectures are not uniformly efficient. 
Static checkpointing is mandatory for Vim-S (Full-FT OOMs without it) and achieves the highest memory reductions (93--95\%), while on MambaVision-T it reduces peak VRAM uniformly across PEFT methods (41--82\%). 
Among zero-shot baselines, DINOv2 (KD-SSL, 0.917 on CIFAR-100) surpasses all fine-tuned models at a fraction of the energy, while autoregressive VLMs remain uncompetitive.

}

}


\let\section\origsection
\bibliographystyle{IEEEtran}
\bibliography{chapter2_references}

\end{document}